\documentclass[sigconf]{acmart}

\usepackage[utf8]{inputenc} % allow utf-8 input
\usepackage[T1]{fontenc}    % use 8-bit T1 fonts
\usepackage{hyperref}       % hyperlinks
\usepackage{url}            % simple URL typesetting
\usepackage{booktabs}       % professional-quality tables
\usepackage{amsfonts}       % blackboard math symbols
\usepackage{nicefrac}       % compact symbols for 1/2, etc.
\usepackage{microtype}      % microtypography
\usepackage{xcolor}         % colors
\usepackage{natbib}
\usepackage{amsthm}
\usepackage[ruled]{algorithm2e}
\usepackage{float}
\usepackage{makecell}
\newtheorem{theorem}{Theorem}

\newtheorem{lemma}{Lemma}

\newtheorem{remark}{Remark}

\usepackage{bm}
\usepackage{subfigure}
\usepackage{colortbl}
\usepackage{enumitem}
\usepackage{framed}
\usepackage{tcolorbox}
\usepackage{setspace}
\usepackage{array}
\usepackage{multirow}

\copyrightyear{2026}
\acmYear{2026}
\setcopyright{cc}
\setcctype{by}
\acmConference[KDD '26]{Proceedings of the 32nd ACM SIGKDD Conference on Knowledge Discovery and Data Mining V.1}{August 09--13, 2026}{Jeju Island, Republic of Korea}
\acmBooktitle{Proceedings of the 32nd ACM SIGKDD Conference on Knowledge Discovery and Data Mining V.1 (KDD '26), August 09--13, 2026, Jeju Island, Republic of Korea}
\acmPrice{}
\acmDOI{10.1145/3770854.3780320}
\acmISBN{979-8-4007-2258-5/2026/08}

\begin{document}

\title{Offline Behavioral Data Selection}

\author{Shiye Lei}
\affiliation{%
  \institution{School of Computer Science \\ The University of Sydney}
  \city{Sydney}
  \country{Australia}
  }
\email{shiye.lei@sydney.edu.au}

\author{Zhihao Cheng}
\affiliation{%
  \institution{School of Computer Science \\ The University of Sydney}
  \city{Sydney}
  \country{Australia}
  }
\email{zhihaocheng111@gmail.com}

\author{Dacheng Tao}
\authornote{Corresponding author.}
\affiliation{%
  \institution{College of Computing \& Data Science\\ Nanyang Technological University}
  %\city{Singopore}
  \country{Singapore}
  }
\email{dacheng.tao@gmail.com}

\renewcommand{\shortauthors}{Shiye Lei, Zhihao Cheng, \& Dacheng Tao}
%%
%% Article type: Research, Review, Discussion, Invited or position
\acmArticleType{Research}
%%
%% Links to code and data
\acmCodeLink{https://github.com/leaveslei/acmart}
%\acmDataLink{htps://zenodo.org/link}
%%
%% Authors' contribution
%\acmContributions{BT and GKMT designed the study; LT, VB, and AP
%  conducted the experiments, BR, HC, CP and JS analyzed the results,
%  JPK developed analytical predictions, all authors participated in
%  writing the manuscript.}
%%
%% Sometimes the addresses are too long to fit on the page.  In this
%% case uncomment the lines below and fill them accodingly.
%%
%% \authorsaddresses{Corresponding author: Ben Trovato,
%% \href{mailto:trovato@corporation.com}{trovato@corporation.com};
%% Institute for Clarity in Documentation, P.O. Box 1212, Dublin,
%% Ohio, USA, 43017-6221}

\begin{abstract}
Behavioral cloning is a widely adopted approach for offline policy learning from expert demonstrations. However, the large scale of offline behavioral datasets often results in computationally intensive training when used in downstream tasks. In this paper, we uncover the striking data saturation in offline behavioral data: policy performance rapidly saturates when trained on a small fraction of the dataset. We attribute this effect to the weak alignment between policy performance and test loss, revealing substantial room for improvement through data selection. To this end, we propose a simple yet effective method, {\it Stepwise Dual Ranking} (SDR), which extracts a compact yet informative subset from large-scale offline behavioral datasets. SDR is build on two key principles: (1) stepwise clip, which prioritizes early-stage data; and (2) dual ranking, which selects samples with both high action-value rank and low state-density rank. Extensive experiments and ablation studies on D4RL benchmarks demonstrate that SDR significantly enhances data selection for offline behavioral data. The code is available at \url{https://github.com/LeavesLei/stepwise_dual_ranking}.
\end{abstract}




\maketitle

\section{Introduction}
\label{sec:intro}

In offline reinforcement learning (RL) \citep{levine2020offline}, where direct interaction with the environment is not feasible, behavioral cloning (BC) offers a simple yet effective supervised approach for learning policies from expert demonstrations \citep{pomerleau1991efficient, bain1995framework}. However, offline behavioral datasets are typically large-scale, limiting the scalability of BC in many downstream tasks. To address this challenge, compressing large datasets into compact yet informative subsets has emerged as a promising direction. These compressed datasets not only facilitate efficient policy fine-tuning across diverse pre-trained models \citep{noh2024efficient}, but also serve as lightweight replay buffers in continual policy learning, effectively mitigating catastrophic forgetting \citep{wolczyk2022disentangling}.

Numerous dataset compression methods, including coreset selection \citep{guo2022deepcore, zhou2022probabilistic} and dataset distillation \citep{wang2018dataset, zhao2021dataset, lei2024comprehensive}, have been developed to construct smaller yet highly representative training sets. However, these approaches are largely tailored to the supervised learning paradigm, which assumes i.i.d. data and access to clean, perfectly labeled samples. In contrast, offline behavioral datasets violate the i.i.d. assumption and exhibit complex interactions between the data distribution and policy performance, which makes the direct application of existing compression techniques nontrivial. Recent works \citep{lei2024offline,lei2025state} apply dataset distillation to synthesize a compact dataset from large-scale offline behavioral data, although the distillation procedure requires computing Hessian matrices and is therefore computationally demanding.

%While various dataset compression techniques, such as coreset selection \citep{guo2022deepcore,zhou2022probabilistic} and dataset distillation \citep{wang2018dataset,zhao2021dataset}, have been proposed to construct smaller but representative training sets for supervised learning, these methods are challenging to apply to offline behavioral data. Unlike standard supervised datasets, offline behavioral data violate the i.i.d. assumption and exhibit more complex dependencies between data distribution and policy performance, making conventional compression approaches ineffective.
\begin{figure*}[t]
\centering
\includegraphics[width=1.\textwidth]{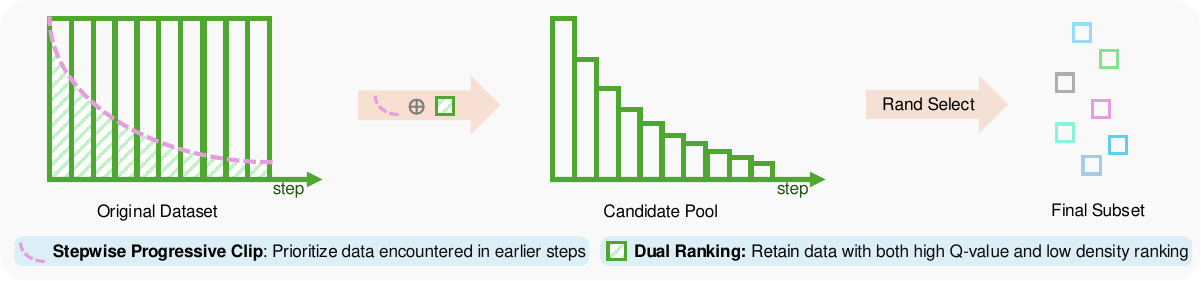}
\vspace{-0.2cm}
\caption{The workflow of Stepwise Dual Ranking. The original dataset is processed through stepwise progressive clip and dual ranking to construct a candidate pool. The final informative subset is then randomly selected from this candidate pool.}
\label{figure:sdr}
%\vspace{-0.3cm}
\end{figure*}

This paper investigates efficient coreset selection for offline behavioral data, aiming to identify a compact yet informative subset from the original behavioral dataset to enable rapid policy training. By analyzing the relationship between policy performance and training sample size, we uncover a striking phenomenon of data saturation in offline behavioral datasets: policy performance rapidly saturates as more training data is added, and a randomly selected subset containing only $1\% \sim 5\%$ of the data can train a policy that achieves $90\%$ of the performance attained using the full dataset. In contrast, supervised image datasets typically require at least $20\%$ of the data to reach similar performance levels. Furthermore, we demonstrate that this data saturation arises from the weak correlation between optimization loss and policy performance. Specifically, reducing the loss does not necessarily translate into meaningful improvements in policy performance, a sharp contrast to conventional supervised learning. % Finally, through theoretical analysis, we show that this data saturation is primarily caused by the state distribution shift between the expert policy and the behavior policy, which limits the effectiveness of traditional optimization objectives in offline BC.

The pronounced data saturation highlights the potential of data selection. Building on a theoretical analysis of per-step allocation and the implicit weighting of behavioral data, we propose a simple yet effective algorithm, \textit{Stepwise Dual Ranking} (SDR), which constructs a compact and informative subset from offline behavioral datasets. Two main strategies are involved in SDR: (1) stepwise progressive clip: prioritize selecting more data visited in the front steps, while choose less data with large steps; and (2) dual ranking: we rank the examples based on density and action values respectively, and prone to select data with both low density rank and high action value ranks. Furthermore, to mitigate the algorithm sensitivity to estimation error, we employ two-stage sampling to generate the final subset. Concretely, we first use SDR to shape a large candidate pool and then randomly select a small set from the candidate pool. This can enhance data diversity and mitigates sensitivity induced by estimation error on data density and action value; please see Figure \ref{figure:sdr} for illustration.

%We first reveal the significant data saturation in behavior cloning, {\it i.e.}, policy performance plateaus despite the inclusion of additional training data. % 为什么会存在严重的data saturation: 1. 和supervised learning一样信息冗余 2. state distribution shift
%We conduct empirical experiments to show the change of performance and training loss. The results demonstrates that policy performance stop to increase with further decreasing the training loss. 
% 会不会是overfitting的影响   
%We then theoretically prove that the significant data saturation is from the state distribution shift between expert policy and behavior policy.
% 详细解释定理

% Algorithm

% Experiments

% and presents the novel finding that policy performance plateaus despite the inclusion of additional training data, indicating the presence of data redundancy within behavioral cloning (BC) datasets.

%BC datasets are constructed based on the D4RL benchmark \citep{fu2020d4rl} through a two-stage process: (1) a decent policy is trained on offline reinforcement learning (RL) datasets using established offline RL algorithms; (2) the actions within these datasets are subsequently corrected using the learned policy.

Extensive experiments on the D4RL benchmark across multiple environments demonstrate that our proposed SDR significantly outperforms the random selection baseline, whereas conventional selection criteria and coreset selection algorithms perform substantially worse in the context of offline behavioral data. 

%\iffalse
Our contributions can be summarized as:
\begin{itemize}[leftmargin=7mm]
    \item We identify and analyze a significant phenomenon of data saturation in offline behavioral datasets, supported by both empirical and theoretical analysis.
    \item We introduce SDR, a novel and efficient approach for constructing an informative subset from large-scale offline behavioral data.
    \item SDR achieves a substantial performance improvement over both traditional selection baselines and existing coreset selection methods.
\end{itemize}
%\fi

\section{Related Works}

%\paragraph{Offline Behavioral Cloning} 
\noindent \textbf{Offline Behavioral Cloning \ \ } Unlike conventional online RL, which requires direct interaction with the environment, offline RL aims to learn policies from a pre-collected dataset, thereby mitigating the costs and risks associated with real-world interactions \citep{levine2020offline}. BC is a widely used offline RL approach that learns policies by mimicking expert demonstrations \citep{pomerleau1991efficient,bain1995framework,torabi2018behavioral}. Since BC directly imitates the expert policy without requiring reward signals, it offers great flexibility and can be applied to a wide range of complex tasks, including autonomous driving \citep{bojarski2016end,ly2020learning} and robotic control \citep{giusti2015machine,niekum2015learning,daftry2016learning}. However, offline behavioral datasets for BC are often large-scale, leading to computational inefficiencies in training and limiting its practicality for downstream applications. This challenge highlights the need for efficient data selection techniques to improve the feasibility of BC in real-world scenarios.

\vspace{2mm}

%\paragraph{Coreset Selection} 
\noindent \textbf{Coreset Selection \ \ } Selecting a representative subset from a large dataset is a promising approach to improving data efficiency in model training \citep{guo2022deepcore}. Coreset selection has been widely applied in various domains, including active learning \citep{park2022active,yehuda2022active}, continual learning \citep{yoon2021online,tiwari2022gcr,hao2023bilevel}, and neural architecture search \citep{shim2021core,prasad2024speeding}. Coreset selection algorithms can be broadly categorized into three main types: (1) Geometry-based methods \citep{welling2009herding,sener2017active}, which aim to preserve the overall data structure in the feature space by selecting a diverse subset; (2) Loss-based methods \citep{toneva2018empirical,paul2021deep}, which prioritize samples with higher training loss, as they are assumed to have greater importance in the learning process; and (3) Gradient-matching methods \citep{mirzasoleiman2020coresets,killamsetty2021grad}, which select a subset that induces similar gradient updates during training compared to using the full dataset. However, these conventional coreset selection techniques face significant challenges when applied to offline behavioral datasets. This is primarily due to (1) the fact that offline behavioral data are collected from various policies, leading to highly heterogeneous and non-i.i.d. data distributions, and (2) the weak and indirect correlation between training loss and policy performance, which makes loss-based selection strategies less effective. As a result, existing coreset selection methods struggle to effectively reduce offline behavioral datasets while maintaining policy performance.

\section{Preliminaries}

\begin{figure*}[t]
\centering
\subfigure[Halfcheetah]{
\begin{minipage}[b]{\textwidth}
\centering
    		\includegraphics[width=0.23\columnwidth]{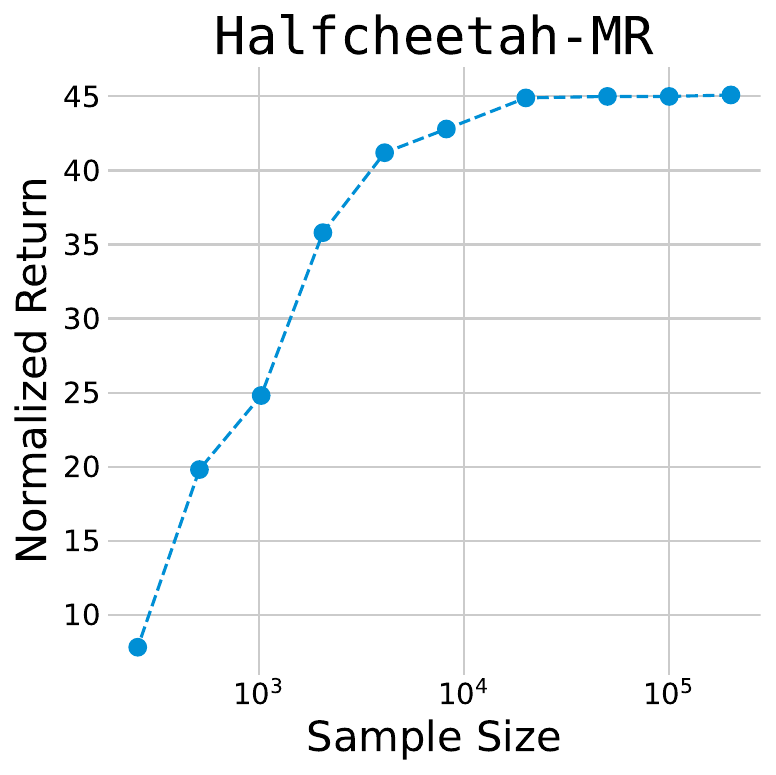}
    		\includegraphics[width=0.23\columnwidth]{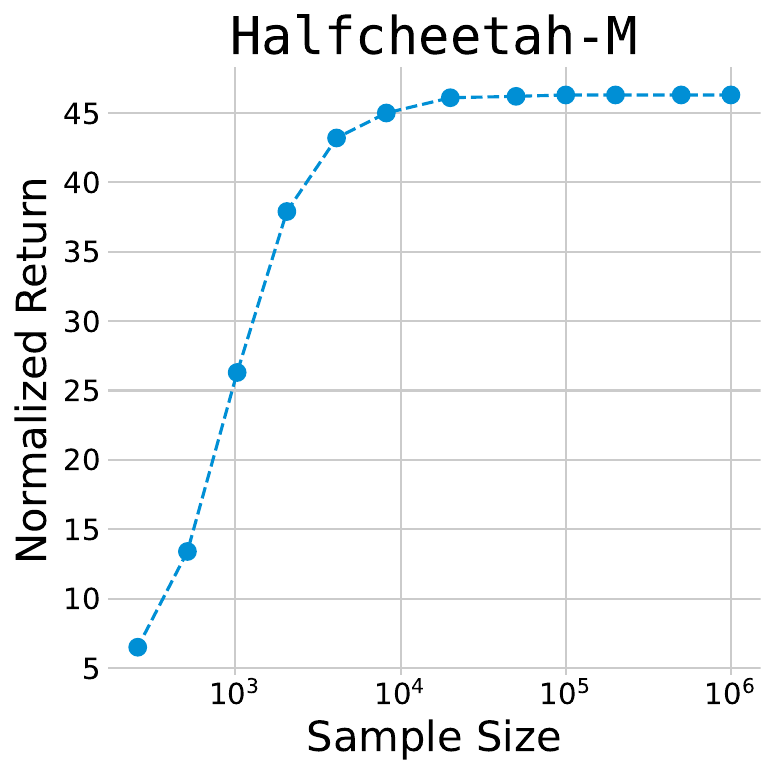}
            \includegraphics[width=0.23\columnwidth]{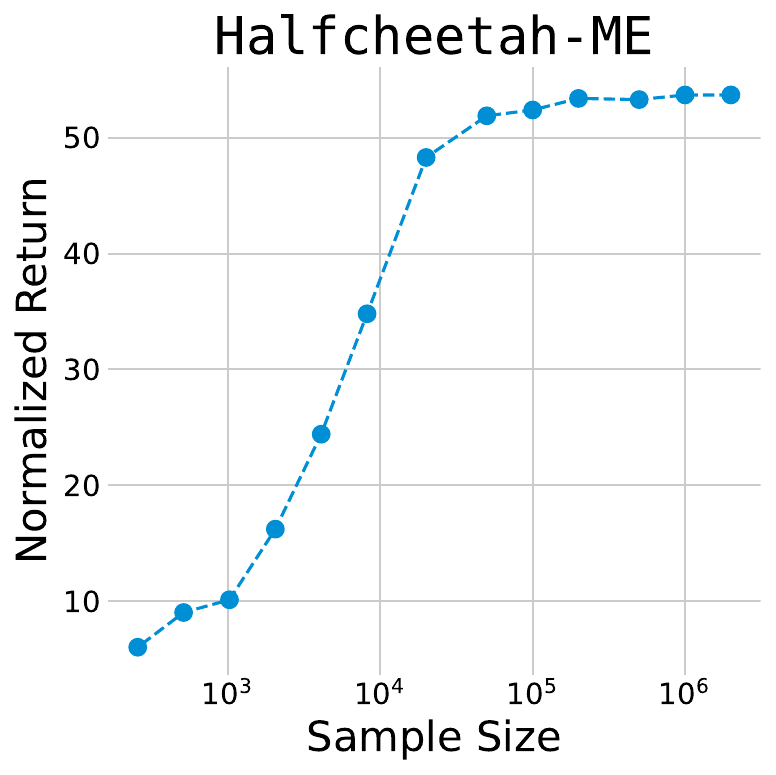}
            \includegraphics[width=0.23\columnwidth]{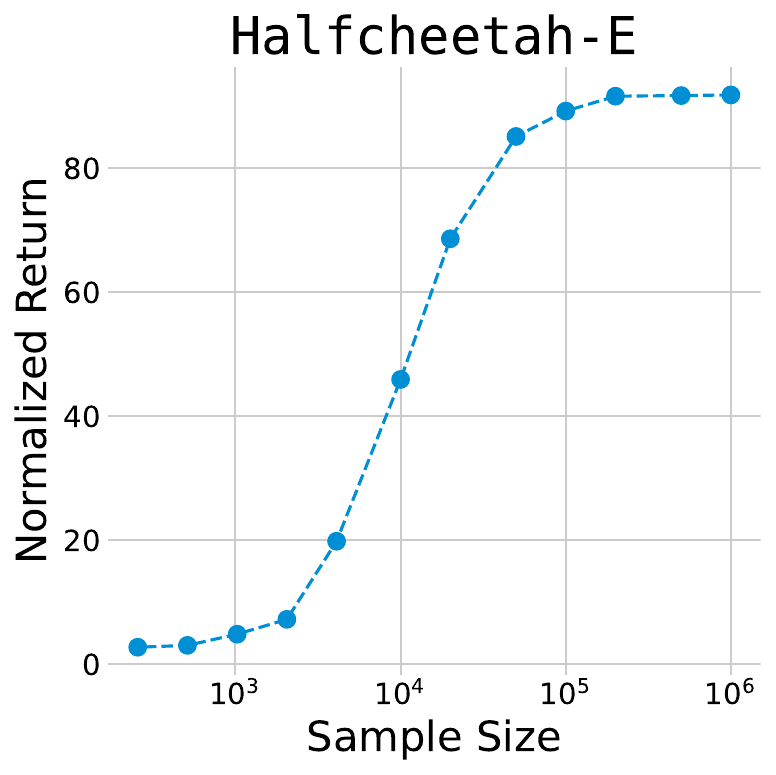}
    		\end{minipage}
		%\label{figure:sample size plot}   
    	}
\subfigure[Hopper]{
\begin{minipage}[b]{\textwidth}
\centering
    		\includegraphics[width=0.23\columnwidth]{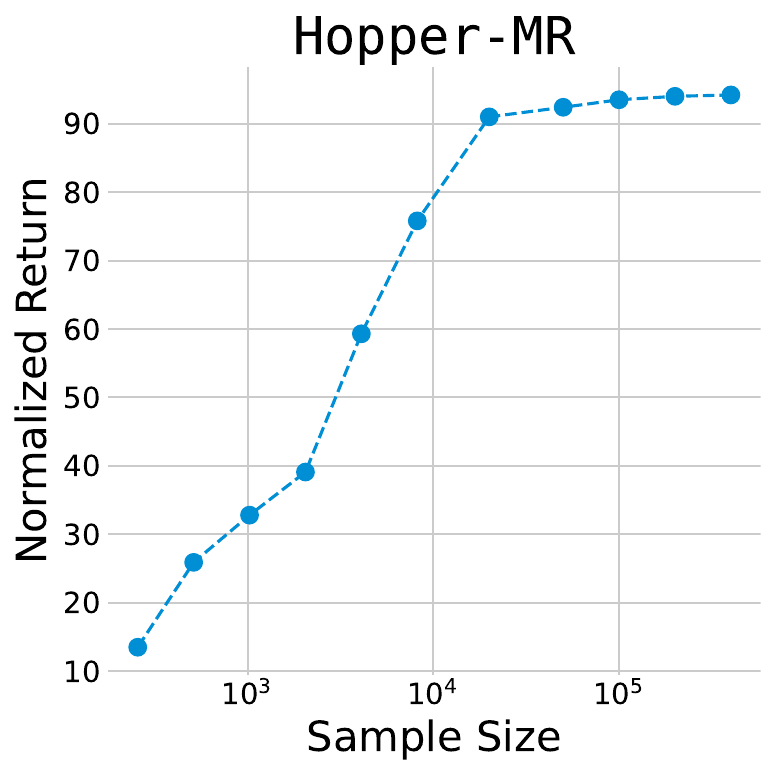}
    		\includegraphics[width=0.23\columnwidth]{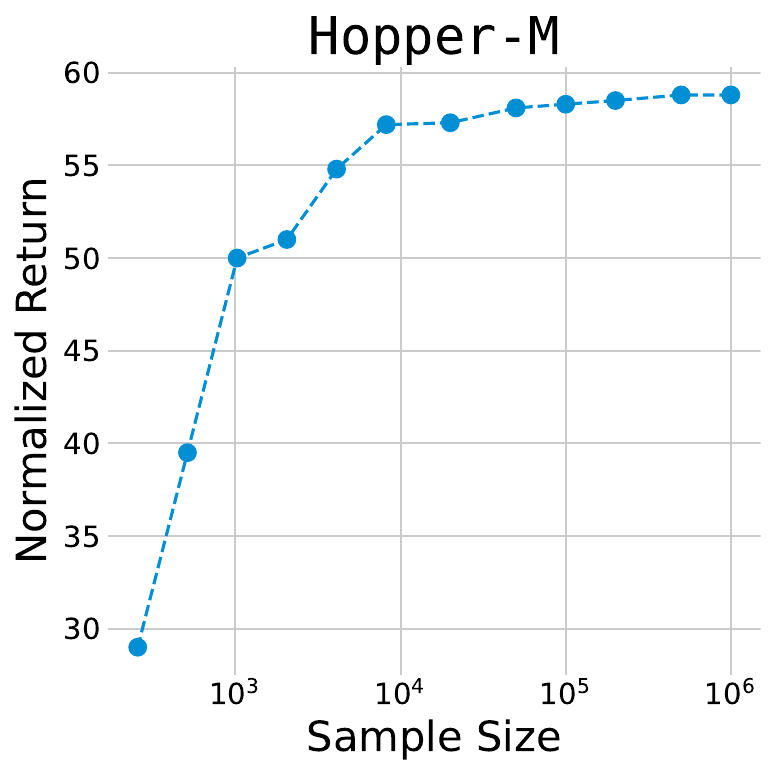}
            \includegraphics[width=0.23\columnwidth]{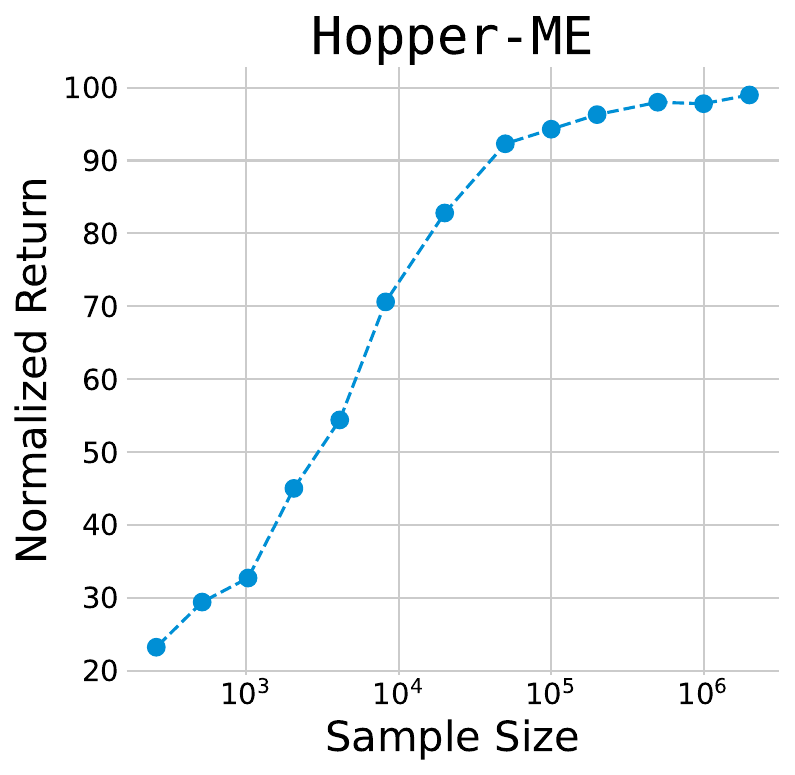}
            \includegraphics[width=0.23\columnwidth]{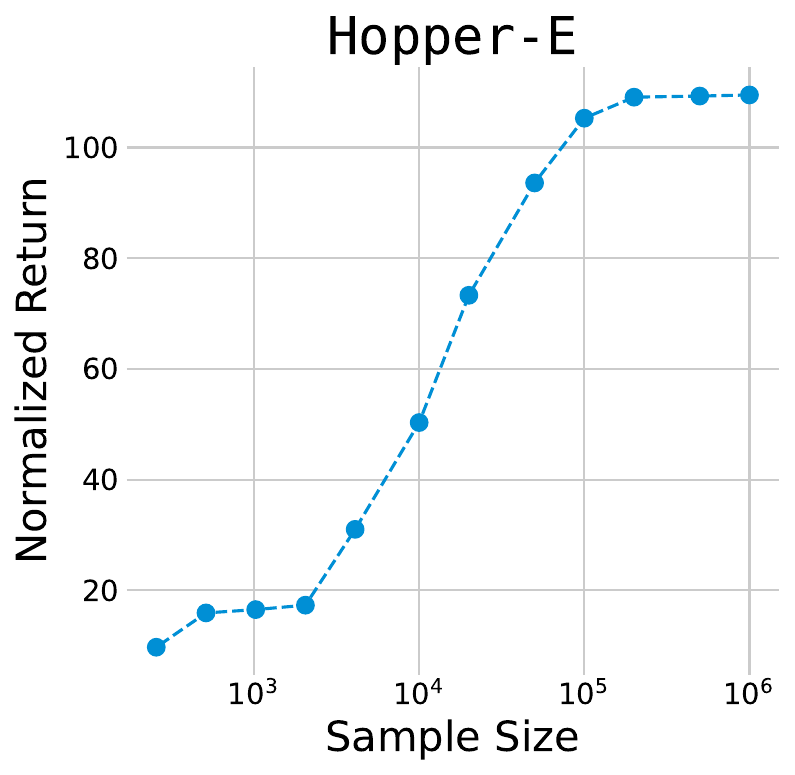}
    		\end{minipage}
		%\label{figure:sample size plot}   
    	}
\subfigure[Walker2D]{
\begin{minipage}[b]{\textwidth}
\centering
    		\includegraphics[width=0.23\columnwidth]{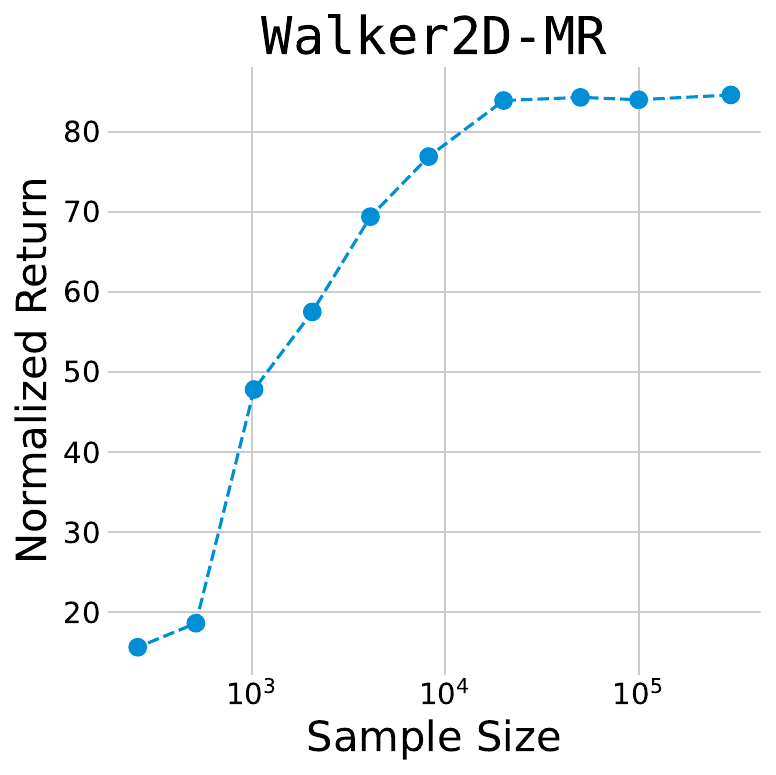}
    		\includegraphics[width=0.23\columnwidth]{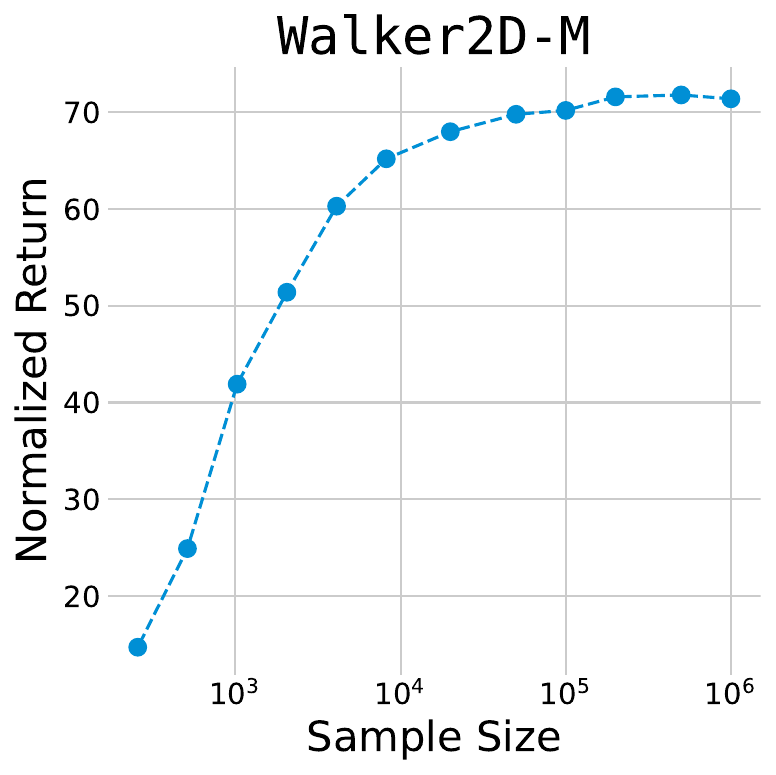}
            \includegraphics[width=0.23\columnwidth]{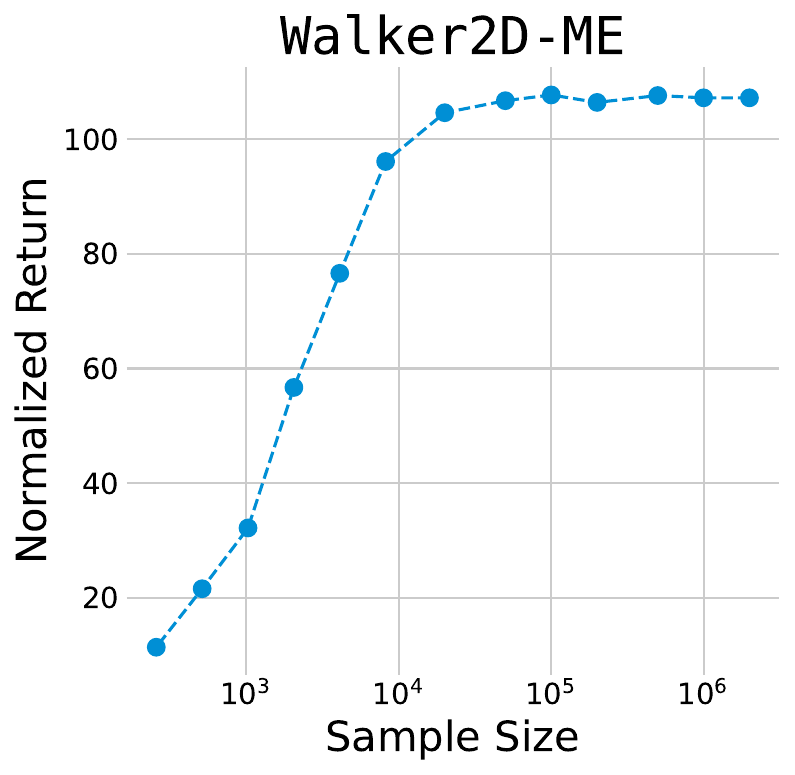}
            \includegraphics[width=0.23\columnwidth]{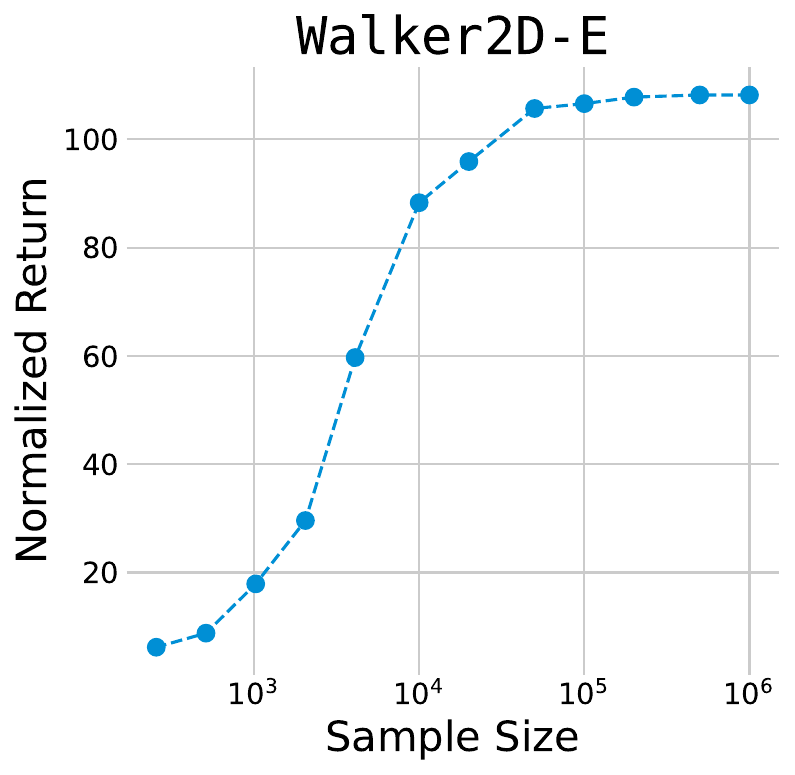}
    		\end{minipage}
		%\label{figure:sample size plot}   
    	}
\vspace{-0.2cm}
\caption{Plots of policy performance (normalized return) as functions of sample size. Each point is averaged over ten trials.}
\label{figure:sample size}
\end{figure*}

\noindent \textbf{Reinforcement Learning} is typically formulated as an episode MDP $\langle\mathcal{S}, \mathcal{A}, \mathcal{T}, r, T, d_0\rangle$, where $\mathcal{S}$ is a set of states ${s} \in \mathcal{S}$, $\mathcal{A}$ is the set of actions ${a} \in \mathcal{A}$, $\mathcal{T}({s}^\prime | {s}, {a})$ denotes the transition probability function, $r ({s}, {a})$ is the reward, $T$ is the horizon length, and $d_0(s)$ is the initial state distribution \citep{sutton2018reinforcement}. $|r(s, a)| \leq 1$ for all $(s,a)\in \mathcal{S}\times \mathcal{A}$. The objective of RL is to learn a policy $\pi$ that maximizes the long-term expected return $J(\pi) = \mathbb{E}_{\pi}\left[\sum_{t=1}^{T} r_t\right]$, where $r_t = r({s}_t, {a}_t)$ is the reward at step $t$. %We define $d_{\pi}^t(s)= \Pr(s_t=s; \pi)$ and $\rho_\pi^t(s, a) = \Pr(s_t=s, a_t=a; \pi)$ as $t$-th step state distribution and state-action distribution, respectively. Then, the average state distribution and state-action distribution of $\pi$ are denoted as $d_{\pi}(s) = \frac{1}{T}\sum_{t=1}^T d_{\pi}^t(s)$ and $\rho_\pi(s, a) =  \frac{1}{T}\sum_{t=1}^T  \rho_\pi^t(s, a) $. 
We define the stepwise state distribution at timestep $t$ as $d_{\pi}^t(s) = \Pr(s_t = s \mid \pi)$, and the average state distribution as $d_{\pi}(s) = \frac{1}{T} \sum_{t=1}^T d_{\pi}^t(s)$. 
%We define $d_{\pi}^t(s)= \Pr(s_t=s; \pi)$ as $t$-th step state distribution, then the average state distribution is denoted as $d_{\pi}(s) = \frac{1}{T}\sum_{t=1}^T d_{\pi}^t(s)$. 
For simplicity, we refer $d_\pi$ as the state distribution of $\pi$. The action-value function of $\pi$ is $q_\pi(s,a) = \mathbb{E}_\pi\left[\sum_{t=1}^{T} r_{t} \mid s_1=s, a_1=a\right]$, which is the expected return starting from $s$, taking the action $a$.

%We primarily use the deterministic form $\pi(s): \mathcal{S} \rightarrow \mathcal{A}$ for simplicity, while the stochastic form $\pi(a|s):\mathcal{S}\times \mathcal{A}\rightarrow [0,1]$ is employed in the theoretical analysis for completeness. 

%We define $d_{\pi}^t(s)= \Pr(s_t=s; \pi)$ and $\rho_\pi^t(s, a) = \Pr(s_t=s, a_t=a; \pi)$ as $t$-th step state distribution and state-action distribution, respectively. Then, the average state visitation and state-action visitation of $\pi$ are denoted as $d_{\pi}(s) = \frac{1}{T}\sum_{t=1}^T d_{\pi}^t(s)$ and $\rho_\pi(s, a) =  \frac{1}{T}\sum_{t=1}^T  \rho_\pi^t(s, a) $. The action-value function of $\pi$ is $q_\pi(s,a) = \mathbb{E}_\pi\left[\sum_{t=1}^{T} r_{t} \mid s_1=s, a_1=a\right]$, which is the expected return starting from $s$, taking the action $a$.

%Without accessing environments, offline RL learns policies from a large offline dataset $\mathcal{D}_\texttt{off}=\{({s}_i,{a}_i,{s}'_i,r_i)\}_{i=1}^{N_\texttt{off}}$ with specially designed Bellman backup. Although $\mathcal{D}_\texttt{off}$ is normally collected by the suboptimal behavior policy, offline RL algorithms can recapitulate a near-optimal policy $\pi^*$ and value function $q_{\pi^*}$ from $\mathcal{D}_\texttt{off}$.

\vspace{1mm}

\noindent \textbf{Behavioral Cloning} is a special offline RL algorithm that learn policy by mimicking high-quality behavioral data. Given the expert demonstrations $\mathcal{D}_\texttt{BC} = \{({s}_i, {a}_i)\}_{i=1}^{N_\texttt{BC}}$, the policy network $\pi_\theta$ parameterized by $\theta$ is trained by cloning the behavior of the expert dataset $\mathcal{D}_\texttt{BC}$ in a supervised manner: $$\min_{\theta}\mathbb{E}_{ \mathcal{D}_\texttt{BC}}[D_{\operatorname{TV}}(\pi_\theta, \hat{\pi}^*)] \coloneq \mathbb{E}_{({s},{a})\sim \mathcal{D}_\texttt{BC}}\left[\sum_{a \in \mathcal{A}}\left|{\pi}_\theta(a \mid s)-\hat{\pi}^*(a \mid s)\right|\right],$$ 
%$\min_{\theta} \ell_\texttt{BC}(\theta, \mathcal{D}_\texttt{BC}) \coloneq \mathbb{E}_{({s},{a})\sim \mathcal{D}_\texttt{BC}}\left[ {\left(\pi_\theta \left({a}|{s}\right) -  \hat{\pi}^*(a|s)\right)^2}\right]$, 
where $\hat{\pi}^*(a|s) = \frac{\sum_{i=1}^{N_\texttt{BC}}\mathbb{I}(s_i=s, a_i=a)}{\sum_{i=1}^{N_\texttt{BC}}\mathbb{I}(s_i=s)}$ is an empirical estimation on $\mathcal{D}_\texttt{BC}$, and $D_{\operatorname{TV}}$ is the total variation distance. %Compared to general offline RL algorithms that deal with subpar $4$-tuples of $\mathcal{D}_\texttt{off}$, BC handles expert $2$-tuples and has better convergence speed.

\vspace{1mm}

\noindent \textbf{Offline Behavior Data \ \ } To construct an offline behavioral dataset consisting of state-action pairs, data is typically collected using various {\it behavior policies} $\pi_\beta$ to gather state information, while actions are labeled either by {\it expert policies} or human annotators. Given a offline RL dataset $\mathcal{D}_\texttt{off}=\{({s}_i,{a}_i,{s}^\prime_i,r_i)\}_{i=1}^{N\texttt{off}}$ collected by suboptimal $\pi_\beta$, where $s_i\sim d_{\pi_\beta}$%and $(s_i,a_i)\sim \rho_{\pi_\beta}$
, a (near) expert policy $\pi^\ast$ can be extracted from $\mathcal{D}_\texttt{off}$ using standard offline RL algorithms. The offline behavioral dataset for BC can then be formulated as $\mathcal{D}=\{(s_i, \pi^\ast(s_i)\}_{i=1}^{N_\texttt{off}}$. This dataset serves as high-quality demonstrations for training RL policies via BC in an offline setting.

\subsection{Offline Behavioral Data Selection}
Given a selection budget of $N$, the objective is to select a subset $\mathcal{S}$ of size $N$ from the original dataset $\mathcal{D}$, such that the policy trained on $\mathcal{S}$ using BC achieves performance comparable to that obtained by training on $\mathcal{D}$. This behavioral data selection problem can be formally defined as:
\begin{align*}
    &\mathcal{S}^\ast = \arg\min_{\mathcal{S}} |J(\mathcal{S}) - J(\mathcal{D})|  \\
    &  \text { s.t. } \quad \mathcal{S}\subset \mathcal{D}, \quad |\mathcal{S}| =N,
\end{align*}
where we slightly abuse notation by using $J(\mathcal{S})$ and $J(\mathcal{D})$ to denote the performance of the policies trained on $\mathcal{S}$ and $\mathcal{D}$, respectively.

\begin{figure*}[t]
\centering
\subfigure[Data Saturation]{
\begin{minipage}[b]{0.245\textwidth}
    \centering
    \includegraphics[width=\columnwidth]{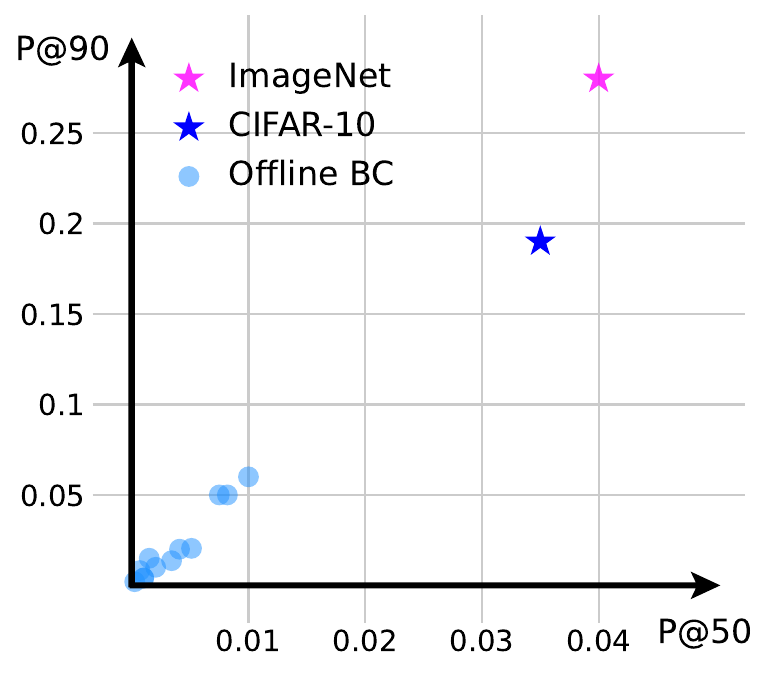}
    %\caption{HalfCheetah}
    \label{figure:data saturation}
\end{minipage}%
}
\hfill
\subfigure[Return vs. Loss]{
\begin{minipage}[b]{0.245\textwidth}
    \centering
    \includegraphics[width=\columnwidth]{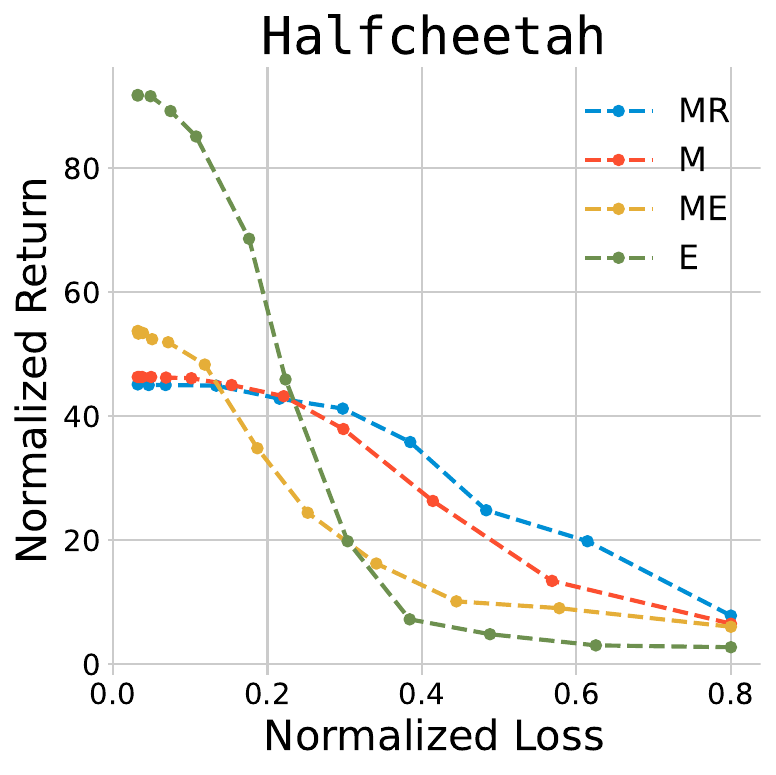}
    %\caption{HalfCheetah}
    % \label{figure:halfcheetah loss}
\end{minipage}%
\hfill
\begin{minipage}[b]{0.245\textwidth}
    \centering
    \includegraphics[width=\columnwidth]{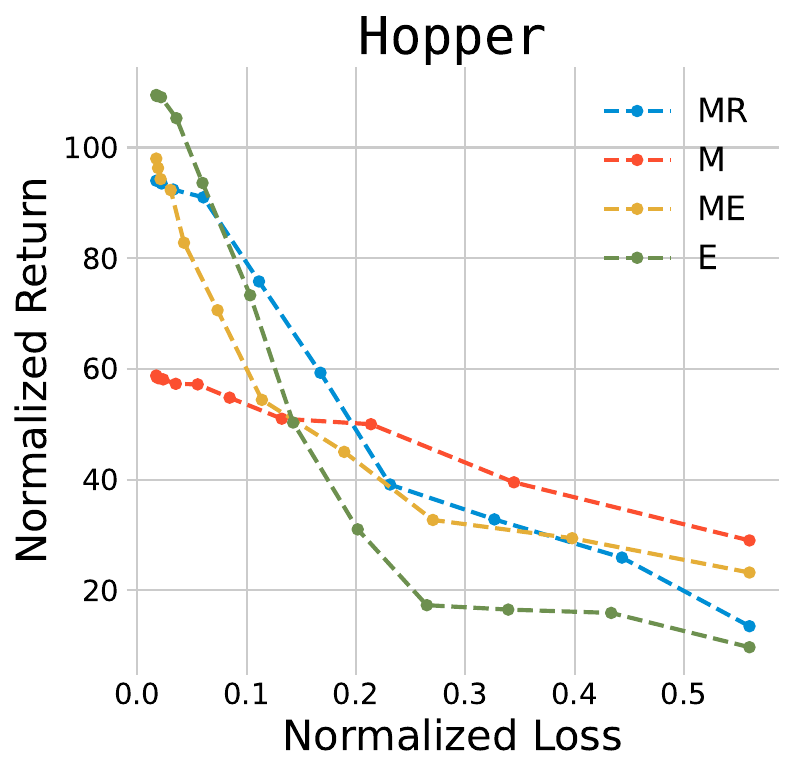}
    % \caption{HalfCheetah}
    % \label{figure:halfcheetah loss}
\end{minipage}%
\hfill
\begin{minipage}[b]{0.245\textwidth}
    \centering
    \includegraphics[width=\columnwidth]{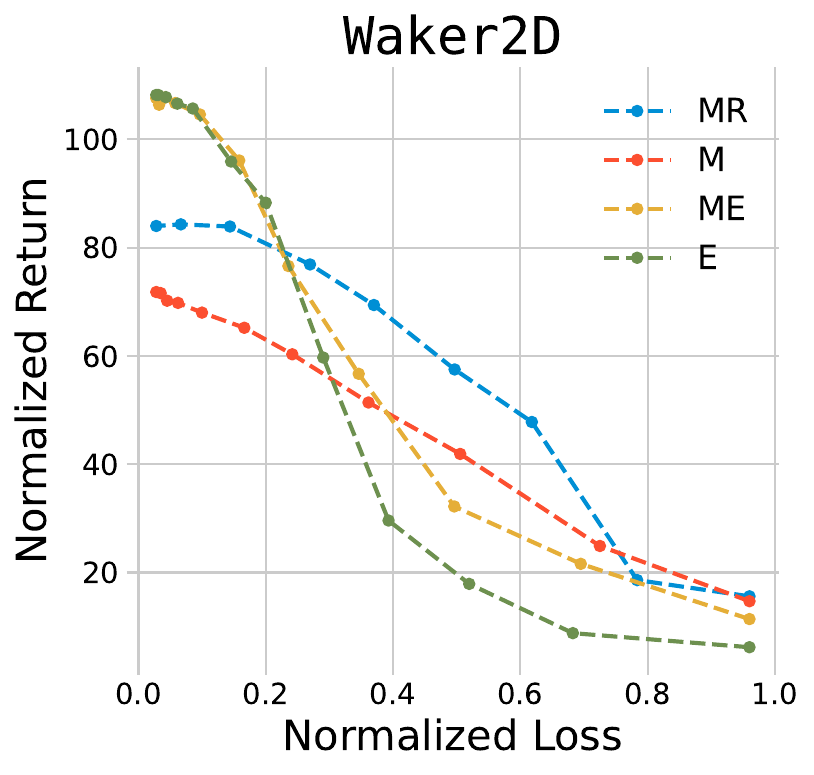}
    % \caption{HalfCheetah}
    % \label{figure:halfcheetah loss}
\end{minipage}%
\label{figure:loss vs return}
}
%\vspace{-2mm}
\caption{(a) $\texttt{P@50}$ and $\texttt{P@90}$ for offline BC datasets and image datasets (ImageNet, CIFAR-10). (b) Plots of policy performance (normalized return) as functions of the loss {\it w.r.t.} real data. Each point is calculated and then averaged over five trials.}
\label{figure: n}
\end{figure*}

\section{Data Saturation in Offline BC Data}
\label{sec:data saturation}

Before introducing our selection algorithm, we begin by investigating the sample efficiency of offline BC datasets, analyzing how policy performance changes with difference sample size. We uncover a surprising phenomenon of data saturation, where increasing the dataset size yields diminishing returns in performance. This phenomenon is explained through both empirical and theoretical analysis.

% behavioral data by how does the policy performance changes with difference sample size. To address this, we begin by presenting the nature of offline RL datasets to provide a clearer context.

\vspace{1mm}

\noindent \textbf{Offline BC Datasets \ \ }  We utilize offline RL data provided by D4RL \citep{fu2020d4rl}, a widely recognized benchmark for building offline BC datasets. Three environments of \texttt{Halfcheetah}, \texttt{Hooper}, \texttt{Walker2D} are utilized for data collection. For each environment, four offline datasets with different qualities are offered by D4RL, {\it i.e.}, \texttt{medium-replay} (\texttt{MR}), \texttt{medium} (\texttt{M}), \texttt{medium-expert} (\texttt{ME}), and \texttt{expert} (\texttt{E}) datasets. Concretely, \texttt{M-R} datasets are generated by a variety of policies, from random to medium-level; \texttt{M} datasets are produced by medium-level policies; \texttt{ME} datasets are collected by policies ranging from medium to expert level, and \texttt{E} datasets are induced by expert policies. For $\texttt{M-R}$, \texttt{M}, and \texttt{ME} those are obtained by suboptimal policies, we can employ advanced offline RL algorithms on them to extract an (near) expert policy. Then we use the obtained policy to relabel actions for states in $\texttt{M-R}$, \texttt{M}, and \texttt{ME}. The relabeled datasets are called offline BC data and can be used for supervised behavioral cloning.

% TODO: 实验细节/步骤
To evaluate the sample efficiency of offline BC datasets, we randomly select a subset from the original dataset and train policies on it. By controlling the size of the selected subset and monitoring policy performance, we analyze the relationship between sample size and policy performance, as illustrated in Figure \ref{figure:sample size}. From the figure, we observe that: (1) policy performance increases as the sample size grows; (2) curves in \texttt{E} saturate more slowly than those in \texttt{MR}, \texttt{M}, and \texttt{ME}; and (3) policy performance gradually plateaus when the sample size exceeds $10,000$, despite the full dataset sizes ranging from hundreds of thousands to millions (with the sample size axis on log scale), particularly in the $\texttt{MR}$ and $\texttt{M}$ scenarios. This plateau indicates that data are extremely saturated in these offline BC datasets. 

To quantify this data saturation, we measure the proportion of data required to train a model or policy that achieves performance close to that obtained with the full dataset. Specifically, we use $\texttt{P@50}$ and $\texttt{P@90}$, which represent the fraction of data needed to reach $50\%$ and $90\%$ of the model performance when trained on the entire dataset, respectively. A smaller number of $\texttt{P@50}$ and $\texttt{P@90}$ means that only a small friction of data is required to train a good model compared to using full datasets, thereby implying severer data saturation. We plot the $\texttt{P@50}$ and $\texttt{P@90}$ for offline BC datasets and two popular image datasets of CIFAR-10 \citep{krizhevsky2009learning} and ImageNet \citep{deng2009imagenet}, as shown in Figure \ref{figure:data saturation}. From the figure we observe that compared to image datasets with $\texttt{P@50}$ of $4\%$ and $\texttt{P@90}$ of $20\%$, these offline BC datasets have surprising low $\texttt{P@50}$ of $1\%$ and $\texttt{P@90}$ of $5\%$, thereby indicating a severe data saturation in offline BC datasets.

%which means only a small friction of data are required to train a good policy compared to using the full data. This indicates that there exists a severe data saturation in offline BC datasets.

%This plateau in offline RL is unexpected, especially when compared to supervised learning settings, where using only around $10\%$ of the data typically results in a significant performance drop \cite{guo2022deepcore}. Therefore, we say that there exists a severe data saturation in offline RL dataset.

% How to quantatively measure data saturation of dataset. 

\subsection{Analysis of Data Saturation}

To better understand the phenomenon of data saturation in offline BC datasets, we analyze the relationship between policy performance and test loss over the full dataset, as shown in Figure \ref{figure:loss vs return}. The figure reveals an interesting trend: in the $\texttt{E}$ setting, minimizing test loss over the full dataset leads to clear improvements in policy performance. However, this correlation weakens considerably in other settings of \texttt{MR}, \texttt{M}, and \texttt{ME}. This observation motivates our hypothesis that data saturation stems from a weak alignment between test loss and policy performance. That is, although increasing the dataset size reduces test loss, it does not necessarily lead to better policy performance. This behavior contrasts with conventional supervised learning, where test loss typically serves as a reliable proxy for model performance.

%{\it i.e.}, increases the normalized return. However, in the $\texttt{MR}$ and $\texttt{M}$ settings, policy performance exhibits a weaker correlation with test loss reduction. Given that the performance curves in \texttt{ME} and \texttt{E} saturate more gradually in Figure \ref{figure:sample size}, we hypothesize that data saturation arises from the weak coupling between policy performance and test loss over the full dataset. In other words, while increasing the sample size effectively reduces test loss, this reduction does not necessarily translate into performance improvement. In contrast, in conventional supervised learning, test loss and model performance are typically well aligned. 

Since \texttt{E} datasets are collected using stronger behavior policies, they exhibit less discrepancy between the state distributions of the behavior and expert policies compared to \texttt{MR}, \texttt{M}, and \texttt{ME} datasets. We theoretically demonstrate that such state distribution shift weakens the alignment between test loss and policy performance, ultimately contributing to data saturation.

%This state distribution shift weakens the correlation between test loss and policy performance, leading to data saturation, as formally analyzed in the following theoretical discussion.

% when further increase sample size and decrease the loss, the policy performance is hardly benefit from xxxx.

% increase sample size, decrease test loss, 但是无法effectively imporve policy performance.

% 用达到final performance 90% 所需要的data的比例来衡量 data saturation。

% TODO: 添加Jupyter notebook中 performance vs. loss 的图

%To further investigate data saturation in BC, we plot the correlation between policy performance and loss over the whole data, as shown in Figure \ref{figure:loss vs return}. The plots demonstrates that when MR and M, the curves are more flat. In other word, increasing the loss cannot bring a ideal improvement w.r.t. policy performance, i.e., expected return. We illustrate this by the following theoretical analysis.

%\subsection{Theoretical Analysis}
%We first present the policy performance bound {\it w.r.t.} BC proposed by \citep{ross2010efficient} as below.
%$\mathbb{E}_{d_{\pi^*}}\left[D_{\operatorname{TV}}\left(\pi, \pi^*\right)\right]$ as below.
\begin{theorem}[\citep{ross2010efficient}]
\label{thm:1}
For policies $\pi^\ast$ and ${\pi}$, we have $
    \left \vert J(\pi) - J(\pi^\ast) \right \vert \leq \epsilon T^2$, where $\epsilon =  \mathbb{E}_{d_{\pi^*}}\left[D_{\operatorname{TV}}\left(\pi, \pi^*\right)\right]$.
\end{theorem}
For $\pi$ trained on offline datasets, Theorem \ref{thm:1} establishes that the policy performance $J(\pi)$, can be non-vacuous bounded by $\mathbb{E}_{d_{\pi^*}}\left[D_{\operatorname{TV}}\left(\pi, \pi^*\right)\right]$ during BC process. However, offline BC datasets are typically collected using suboptimal behavioral policy $\pi_\beta$ rather than the expert policy $\pi^\ast$. To this end, during the training on offline behavioral datasets, optimization is performed on $\mathbb{E}_{d_{\pi_\beta}}\left[D_{\operatorname{TV}}\left(\pi, \pi^*\right)\right]$ instead of $\mathbb{E}_{d_{\pi^*}}\left[D_{\operatorname{TV}}\left(\pi, \pi^*\right)\right]$. To demonstrate the influence caused by the discrepancy between $d_{\pi_\beta}$ and $d_{\pi^*}$, we propose the following theorem.
%The following theorem quantifies the influence caused by the discrepancy between $d_{\pi_\beta}$ and $d_{\pi^*}$.
\begin{theorem}
\label{thm:main}
For two policies $\pi^\ast$ and ${\pi}$, we have
\begin{align*}
& \left|\mathbb{E}_{d_{\pi_\beta}}\left[D_{\operatorname{TV}}\left(\pi, \pi^*\right)\right]-\mathbb{E}_{d_{\pi^*}}\left[D_{\operatorname{TV}}\left(\pi, \pi^*\right)\right]\right| \nonumber \\
& \qquad \qquad \qquad\leq 2{D}_{\operatorname{TV}}\left(d_{\pi_\beta}, d_{\pi^*}\right) \sup _s D_{\operatorname{TV}}\left(\pi, \pi^*\right) . \nonumber
\end{align*}
\end{theorem}
\begin{proof}
We begin by rewriting the left-hand side (LHS) as an integral over states:
\begin{align}
    \text{LHS} & = \left \vert \int\left(d_{\pi_\beta}(s)-d_{\pi^*}(s)\right) \operatorname{KL}\left(\pi \| \pi^*\right) \text{d}s \right \vert \nonumber \\
    & \leq  \int\left\vert d_{\pi_\beta}(s)-d_{\pi^*}(s)\right\vert \operatorname{KL}\left(\pi \| \pi^*\right) \text{d} s  \nonumber \\
    & \leq 2 \operatorname{D}_\text{TV}\left(d_{\pi_\beta}, d_{\pi^*}\right) \sup _s \operatorname{KL}\left(\pi \| \pi^*\right) \nonumber
\end{align}
The first inequality follows from the triangle inequality for integrals, and the second uses the definition of total variation distance and the fact that $\operatorname{KL}(\pi \| \pi^*)$ is uniformly bounded by its supremum. This concludes the proof.
\end{proof}
%The proof detail can be found in Appendix \ref{app:thm}. 
According to Theorem \ref{thm:main}, a larger state distribution shift, measured by  ${D}_\text{TV}\left(d_{\pi_\beta}, d_{\pi^*}\right)$, results in a greater gap between the practical loss $\mathbb{E}_{d_{\pi_\beta}}\left[D_{\operatorname{TV}}\left(\pi, \pi^*\right)\right]$ and true loss $\mathbb{E}_{d_{\pi^*}}\left[D_{\operatorname{TV}}\left(\pi, \pi^*\right)\right]$. This result theoretically explains how state distribution shift induces an optimization gap, leading to the phenomenon where further reducing the test loss by increasing sample size does not necessarily reduce the true loss. Consequently, this prevents meaningful policy improvement and results in data saturation in offline behavioral datasets. %Furthermore, this analysis explains why \texttt{ME} and \texttt{E} exhibit less severe data saturation compared to \texttt{MR} and \texttt{M}: the former are collected by stronger behavior policies, which exhibit a smaller distribution shift from the expert policy. 

\section{Algorithm}
\label{sec:algorithm}

The severe data saturation in offline behavioral data underscores the potential for data selection. In this section we design an effective data selection algorithm that identifies a subset of informative state-action pairs for rapid policy training. Given the original offline dataset $\mathcal{D}=\{\mathcal{D}_0, \mathcal{D}_{1},\cdots,\mathcal{D}_{T-1}\}$, where $\mathcal{D}_t=\{s_i,\pi^\ast(s_i)\}_{i=1}^{|\mathcal{D}_t|}$ consists of states collected by the behavioral policy $\pi_\beta$ at timestep $t$, we construct a subset $\mathcal{S}=\{\mathcal{S}_0,\mathcal{S}_1, \cdots, \mathcal{S}_{T-1}\}$ such that $\mathcal{S}_t\subset\mathcal{D}_t$, and $|\mathcal{S}_t|=n_t$. Our analysis proceeds in two steps: (1) examining the impact of the per-step selection size $n_t$ on $J(\mathcal{S})$, {\it i.e.}, how many samples should be selected at each step; and (2) determining which specific examples should be chosen given $\mathcal{D}_t$. For simplicity, we denote $d_{\pi^\ast}$ and $d_{\pi_\beta}$ as $d_\ast$ and $d_\beta$, respectively. 

\subsection{Theoretical Analysis}
The variant of Theorem \ref{thm:1} offers a tighter bound by decomposing $\epsilon$ into stepwise errors as below.
\begin{lemma}[\citep{ross2010efficient}]
\label{lemma:step error}
    For policies $\pi^\ast$ and $\pi$, let $\epsilon_t = \mathbb{E}_{s \sim d_{*}^t}\left[D_{\operatorname{TV}}\left(\pi, \pi^\ast\right)\right]$,  we have $\vert J(\pi) - J(\pi^\ast) \vert \leq \sum_{t=0}^{T-1} (T-t) \epsilon_t$.
\end{lemma}
According to Lemma \ref{lemma:step error}, the error incurred in the earlier steps, \textit{i.e.}, $\epsilon_t$ for small $t$, is more critical to policy performance $J(\pi)$. Since $s \sim d_\beta^t$ for $s\in \mathcal{D}_t$, selecting from $\mathcal{D}_t$ can be viewed as sampling $n_t$ examples from $d_\beta^t$ to minimize the stepwise error $\epsilon_t$ evaluated under $d_*^t$. Consequently, we derive a bound to quantify the gap between the expected error $\epsilon_t$ and its empirical estimate $\hat{\epsilon_t}$.
\begin{lemma}
\label{lemma:generalization}
Suppose $m$ examples are i.i.d. sampled from $d'$, then $\pi$ is trained on them and test on the target distribution $d$. Then the expected error $\epsilon=\mathbb{E}_{d}\left[D_{\operatorname{TV}}\left(\pi, \pi^\ast\right)\right]=\mathbb{E}_{d'}\left[\frac{d(s)}{d'(s)}D_{\operatorname{TV}}\left(\pi, \pi^\ast\right)\right]$, and empirical error $\hat{\epsilon}=\frac{1}{m}\sum_{i=1}^m \frac{d(s_i)}{d'(s_i)}D_{\operatorname{TV}}\left(\pi, \pi^\ast\right)$. Let $w_i=\frac{d(s_i)}{d'(s_i)}$. Then, with the probability of at least $1-\delta$, the following inequality holds:% for all $\pi\in \Pi$:
    \begin{equation*}
        \epsilon \leq \hat{\epsilon} + \sqrt{\frac{\sum_{i=1}^m w^2_i \log\frac{2}{\delta}}{m^2}}.
    \end{equation*}
\end{lemma}
we observe that $\frac{1}{m}\sum_{i=1}^m w^2_i$ represents an empirical estimation in terms of $\mathcal{C}=\mathbb{E}_s\left[\left({d\left(s\right)}/{d'\left(s\right)}\right)^2\right]$, where $\mathcal{C}=\chi^2(d\|d') + 1$, and can be interpreted as the distance between $d$ and $d'$. The empirical error $\hat{\epsilon}$ can be typically optimized to zero during the training process, we obtain the bound $\epsilon \lesssim  \sqrt{\frac{\mathcal{C}_t  \log\frac{2}{\delta}}{n_t}}$. Then, combining Lemmas \ref{lemma:step error} and \ref{lemma:generalization} leads the following bound.
\begin{theorem}
\label{thm:diffbound}
Given the original dataset $\mathcal{D}$, where $s\in \mathcal{D}$ is sampled from $d_\beta$, and treating the policy trained on $\mathcal{D}$ as expert $\pi^\ast$. Then we randomly select $n_t$ examples from $\mathcal{D}$ for each step to constitute the subset $\mathcal{S}$. Let $\mathcal{C}_t=\mathbb{E}_s\left[\left(\frac{d_\ast^t(s)}{d_\beta^t(s)}\right)^2\right]$. If policy $\pi$ is trained on $\mathcal{S}$ with zero training error, then with the probability of at least $(1-\delta)^T$, the following inequality holds:
    \begin{equation*}
        |J(\mathcal{S}) - J(\mathcal{D})| \lesssim \sum_{t=0}^{T-1} (T-t) \sqrt{\frac{\mathcal{C}_t \log\frac{2}{\delta}}{n_t}}.
    \end{equation*}
\end{theorem}
\begin{remark}
    As shown in Theorem \ref{thm:diffbound}, the performance gap between policies trained on $\mathcal{S}$ and $\mathcal{D}$ is related to $\sqrt{{\mathcal{C}_t}/{n_t}}$. Typically, $d_{\beta}^t$ gradually diverges from $d_{\ast}^t$ due to the cascade error, causing $\mathcal{C}_t$ to increase with larger $t$. This highlighting the importance of data sampling in the later timesteps.
\end{remark}
Thus, the problem of behavioral data selection can be framed as the following optimization problem:
\begin{align*}
   & n^\ast_0,  n^\ast_1, \cdots,n^\ast_{T-1} = \arg\min_{n_t}\sum_{t=1}^T (T-t) \sqrt{\frac{\mathcal{C}_t}{n_t}} \nonumber \\
   & \qquad \qquad \text { s.t. } \quad \sum_{i=0}^{T-1} n_t=N
\end{align*}
Solving the above optimization problem yields $n_t \propto \sqrt[3]{(T-t)^2 \mathcal{C}_t}$. Since the distributional divergence $\mathcal{C}_t$ is typically difficult to estimate in the offline setting, directly computing $\sqrt[3]{(T - t )^2 \mathcal{C}_t}$ is not practically feasible. Therefore, we focus on analyzing the general characteristics of $n_t$ for manually constructing a reasonable functional form. A conservative yet practical assumption is that $n_t$ is primarily influenced by $(T - t )$, suggesting that (1) {\it $n_t$ decreases monotonically with $t$}. On the other hand, since $\mathcal{C}_t$ tends to increase over time, (2) {\it the rate at which $n_t$ decreases should gradually slow down}.

After characterizing the relationship between $n_t$ and the timestep $t$, we proceed to investigate how to choose data from each $\mathcal{D}_t$. Since offline data are sampled from $d_\beta^t$, while the stepwise error $\epsilon_t$ is evaluated under $d_*^t$, we estimate the empirical error via importance sampling as $\hat{\epsilon}_t=\mathbb{E}_{s\sim \mathcal{D}_t}\left[\frac{d_\ast^t(s)}{d_\beta^t(s)}D_{\operatorname{TV}}(\pi,\pi^\ast)\right]$. This formulation highlights that the ratio $d_\ast^t/d_\beta^t$ serves as an importance weight. Consequently, we prioritize samples with higher values of this ratio when selecting from $\mathcal{D}_t$.

Calculating ${d_{\ast}^t(s)}/{d_{\beta}^t(s)}$ requires access to the stepwise density $d_{\pi^\ast}^t(s)$ and $d_{\beta}^t(s)$. For $d_{\beta}^t(s)$, although it can be directly estimated based on $\mathcal{D}_\texttt{off}^t$, we instead approximate it with $d_{\beta}(s)$ estimated from the entire $\mathcal{D}_\texttt{off}$ for computational efficiency. For $d_{\ast}$, since we do not have access to the environment, we approximate it using $q_{\pi^\ast}\left(s,\pi^\ast\left(s\right)\right)$, under the intuition that states with higher action values are more likely to be visited by the expert policy $\pi^\ast$ to achieve higher returns. Importantly, our algorithm only requires relative rankings rather than exact density values. Therefore, we use $q_{\pi^\ast}$ and $d_{\ast}$ as a coarse estimations of $d_{\ast}^t$ and $d_{\beta}^t$, respectively.

\begin{table*}[t]
\scriptsize
    \centering
    \caption{Data selection performance on D4RL offline datasets with the budget of $1024$. The results are averaged across five seeds. The best selection result for each dataset is marked with bold scores.}
    %\vspace{-0.1cm}
    \resizebox{1\linewidth}{!}{
    \begin{tabular}{c  ccc  ccc  ccc c}
    \toprule
     \multirow{2}{*}{Method}  & \multicolumn{3}{c}{Halfcheetach} & \multicolumn{3}{c}{Hopper} & \multicolumn{3}{c}{Walker2D} & \multirow{2}{*}{Average} \\
     & M-R & M & M-E & M-R & M & M-E & M-R & M & M-E \\
     %\specialrule{1pt}{0.2\jot}{0.2pc}
     \midrule
     Random & $28.0$ & $26.3$ & $10.1$ & $32.8$ & $50.0$ & $32.7$ & $47.8$ & $41.9$ & $32.2$ & $27.0$   \\
     Top Reward & $10.6$ & $1.4$ & $2.2$ & $5.9$ & $6.9$ & $3.9$ & $3.6$ & $2.7$ & $7.6$ & $5.0$   \\
     Top Q-value &  $4.4$ & $2.3$ & $1.7$ & $6.1$ & $7.7$ & $9.1$ & $3.8$ & $13.1$ & $4.3$ & $5.8$   \\
     \specialrule{0.2pt}{0.2\jot}{0.2pc}
     Herding  & $0.7$ & $1.0$ & {$3.3$} & $0.6$ & {$4.4$} & $2.9$ & $3.6$ & $2.7$ & $1.1$   & $2.3$ \\
     GraNd  & {$2.0$} & $2.1$ & $3.1$ & {$6.4$} & $14.5$ & $1.3$ & {$2.7$} & $5.9$ & $3.3$   & $4.6$ \\
     GradMatch  & {{$3.3$}} & {$6.8$} & $\bm{18.9}$ & {{$29.3$}} & $18.3$ & {$43.7$} & {{$23.9$}} & {$13.6$} & {$44.5$}   & {$22.5$}\\
     %\specialrule{1pt}{0.2\jot}{0.2pc}
     \midrule
     SDR (ours) & {$\bm{33.9}$} & $\bm{28.2}$ & $10.5$ & $\bm{41.6}$ & $\bm{54.0}$ & $\bm{37.4}$ & $\bm{57.0}$ & $\bm{45.9}$ & $\bm{48.4}$ & $\bm{40.8}$   \\
    \bottomrule
    \end{tabular}
    }
    \label{tab:selection result}
\end{table*}

\begin{algorithm}[t]
\SetKwInOut{KwIn}{Input}
\SetKwInOut{KwOut}{Output}
\caption{Stepwise Dual Ranking}
\label{alg:obd}
\KwIn{Offline dataset $\mathcal{D}_\texttt{off}$, selection budget $N$, trajectory length $T$, quantile function $\mathcal{F}$}
\KwOut{Selected subset $\mathcal{S}$}

$\pi^*, q_{\pi^*} \leftarrow \texttt{OfflineRL}(\mathcal{D}_\texttt{off})$ \\
Construct dataset $\mathcal{D} = \{(s_i, \pi^*(s_i))\}_{i=1}^{N_\texttt{off}}$ for $s_i \in \mathcal{D}_\texttt{off}$ \\
$d_\beta(s) \leftarrow \texttt{DensityEstimation}(\mathcal{D})$ \\

\textbf{Candidate Pool Construction} \\
Initialize candidate pool $\mathcal{P} \leftarrow \emptyset$ \\

\For{$t = 1$ \textbf{to} $T$}{
    $\alpha_t \leftarrow \mathcal{F}(t)$ \\
    $q_t \leftarrow$ the $\alpha_t$-quantile of $\{q_{\pi^*}(s,a) \mid (s,a) \in \mathcal{D}_t\}$ \\
    $d_t \leftarrow$ the $(1 - \alpha_t)$-quantile of $\{d_\beta(s) \mid (s,a) \in \mathcal{D}_t\}$ \\
    
    \For{$(s,a) \in \mathcal{D}_t$}{
        \If{$q_{\pi^*}(s,a) \geq q_t$ \textbf{and} $d_\beta(s) \leq d_t$}{
            Add $(s, \pi^*(s))$ to $\mathcal{P}$
        }
    }
}
$\mathcal{S} \leftarrow$ randomly select $N$ samples from $\mathcal{P}$ \\
\Return $\mathcal{S}$
\end{algorithm}

\subsection{Stepwise Dual Ranking}
With the aforementioned analysis, we propose two key strategies, stepwise progressive clip and dual ranking, to effectively select informative state-action pairs for policy training.

\vspace{1mm}

\noindent \textbf{Stepwise Progressive Clip  \ \ }
According to the theoretical analysis {\it w.r.t.} $n_t$, we prioritize sampling more data from $\mathcal{D}_\texttt{off}^t$ when $t$ is small, but gradually reducing the sampling size in a more moderate rate as $t$ increases, {\it i.e.}, $n_{t} - n_{t+1}$ decreases as $t$ increase.

%We prioritize sampling more data from $\mathcal{D}_\texttt{off}^t$ when $t$ is small, gradually reducing the sampling rate for $\mathcal{D}_\texttt{off}^t$ as $t$ increases. This strategy is motivated by the fact that, for larger $t$, the distribution of the weight $d_{\pi^\ast}^t(s)/d_{\pi_\beta}^t(s)$ becomes more uneven, with examples in $\mathcal{D}_\texttt{off}^t$ exhibit with higher weights. To this end, fewer top examples with high weights are required to accurately estimate the stepwise loss $\mathbb{E}_{\mathcal{D}_\texttt{off}^t}\left[{d_{\pi^\ast}^t(s)}/{d_{\pi_\beta}^t(s)}D_{\operatorname{TV}}\left(\pi, \pi^*\right)\right]$. 

% We prioritize sampling more data from $\mathcal{D}_\texttt{off}^t$ when $t$ is small, while reducing the sampling rate for larger $t$. This ensures that early-stage data, which generally exhibit lower distributional shift, are well represented.

% 讨论使用除法时会有较大的 estimation variance 

\vspace{1mm}

\noindent \textbf{Dual Ranking \ \ }A rational approach for selecting $(s,a)$ pairs from $\mathcal{D}_\texttt{off}^t$ is to prioritize those with a high estimated weight $q_{\pi^\ast}(s,a)/d_{\pi_\beta}(s)$. However, two sources of estimation error must be considered: (1) we employ $q_{\pi^\ast}$ and $d_{\beta}$ to approximate $d_{\ast}^t$ and $d_{\beta}^t$, respectively, introducing approximation errors; (2) inherent estimation noise exists in evaluating $q_{\pi^\ast}$ and $d_{\beta}$. Furthermore, these two sources of error are compounded and further amplified by the division operation $q_{\pi^\ast}/d_{\beta}$. To enhance robustness against these compounding errors, we propose a dual ranking selection strategy. Specifically, we independently rank data points in descending order based on $q_{\pi^\ast}$ and in ascending order based on $d_{\beta}$. We then prioritize selecting pairs that achieve high ranks {\it in both criteria}. This dual ranking approach mitigates reliance on exact numerical values and avoids the division operation, thereby improving robustness against inaccuracies due to using   $q_{\pi^\ast}$ and $d_{\beta}$.

By integrating the two strategies above, we formulate {\it {S}tepwise Dual Ranking} (SDR), which serves as the core of our algorithm: (1) we use a predefined monotonic function $\mathcal{F}:t\rightarrow [0,1]$ to control the sampling size $n_t$ by $\alpha_t=\mathcal{F}(t)$; and (2) for each step-specific original dataset $\mathcal{D}_t$, we rank its elements based on both $q_{\pi^\ast}$ and $d_{\beta}$. We then select examples that simultaneously lin in the  top-$\alpha_t$ percentile {\it w.r.t.} $q_{\pi^\ast}$ and the bottom-$(1-\alpha_t)$ percentile {\it w.r.t.} $d_{\beta}$.

% We typically employ a predefined monotonic function, $\texttt{QuantileFunc}$ (e.g. $\frac{t}{T}$), to determine the proportion of data to be sampled from $\mathcal{D}_\texttt{off}^t$. Concretely, for a given step $t$, we compute the quantile threshold as $\alpha = \texttt{QuantileFunc}(t)$. Then we select all $(s,a)$ pairs in $\mathcal{D}_\texttt{off}^t$ that satisfy $\mathcal{F}_{\mathcal{D}_\texttt{off}^t}(d_{\pi_\beta}(s)) \leq 1 - \alpha$ and $\mathcal{F}_{\mathcal{D}_\texttt{off}^t}(q_{\pi^\ast}(s,a)) \geq \alpha$, where $\mathcal{F}_{\mathcal{D}_\texttt{off}^t}(d_{\pi_\beta}(s))$ and $\mathcal{F}_{\mathcal{D}_\texttt{off}^t}(q_{\pi^\ast}(s,a))$ denote the respective quantile values of $d_{\pi_\beta}(s)$ and $q_{\pi^\ast}(s,a)$ in $\mathcal{D}_\texttt{off}^t$.

% 把stepwise clip和dual ranking合起来解释
%When combine these two criteria, we derive the stepwise dual ranking, which is the core of our algorithm.
% Additionally, using a pre-designed quantile function, such as $\texttt{QuantileFunc}(t) = \frac{t}{T}$, we select all $(s,a)$ pairs in $\mathcal{D}_\texttt{off}^t$ that satisfy $p_d(s) \leq 1 - \alpha$ and $p_q(s) \geq \alpha$, where $p_d(s)$ and $p_q(s)$ represent the respective quantiles of $d(s)$ and $q(s,a)$.

\vspace{1mm}
    
\noindent \textbf{Two-stage Sampling \ \ }
Since the selected subset is typically much smaller than the original dataset, directly applying stepwise dual ranking makes the selection process highly sensitive to the estimated value of $d_{\beta}$ and $q_{\pi^\ast}$. To improve robustness, we adopt a two-stage sampling strategy. First, we apply stepwise dual ranking to construct a candidate pool, whose size lies between $|\mathcal{S}|$ and $|\mathcal{D}|$. This ensures that the most informative samples are retained while maintaining diversity. Then, we randomly sample a subset from this candidate pool to form the final coreset. 
%By leveraging this two-stage approach, we first use stepwise ranking to generate a coarse but sufficiently large intermediate selection, followed by random sampling to refine the coreset. 
This strategy enhances data diversity and mitigates the sensitivity to aforementioned estimation error. The pseudo-code for SDR is in Algorithm \ref{alg:obd}.

\section{Experiments}
\label{sec:exp}
We evaluate our SDR on offline behavioral datasets based on D4RL, with comparisons focused on typical selection strategies and diverse coreset selection methods. 
%random selection, considering the previous coreset selection algorithms are much worse than the plain random selection.
Moreover, we conduct serious ablation study to show the effectiveness of the strategies in SDR.
% 实验说明传统的coreset selection在BC中不work，甚至远不如random selection

% 分析传统core select的缺陷

% 我们的方法

%\noindent \textbf{Setup\ \ } We adopt Cal-QL~\citep{nakamoto2023calql}, a state-of-the-art offline RL algorithm, to extract high-quality expert policy $\pi^*$ and value estimates $q_{\pi^*}$ from the offline dataset $\mathcal{D}_\texttt{off}$. To determine the stepwise quantile threshold, we use the function $\mathcal{F}(t) = \lambda \cdot \tanh\left(\frac{t}{100}\right)$, with the hyperparameter $\lambda$ set to $0.2$. Policy networks are implemented as 4-layer MLPs with a hidden width of 256. We train using a batch size of $256$, a learning rate of $3 \times 10^{-4}$, and the Adam optimizer~\citep{kingma2014adam}. For experiments on selected subsets, we repeat the data to match the size of the original dataset and train for 20 epochs to ensure equal training time for fair comparison.

\vspace{1mm}
\noindent \textbf{Setup\ \ } We adopt Cal-QL~\citep{nakamoto2023calql}, a state-of-the-art offline RL algorithm, to extract high-quality expert policy $\pi^*$ and value estimates $q_{\pi^*}$ from the offline dataset $\mathcal{D}_\texttt{off}$. To determine the stepwise quantile threshold, we use the function $\mathcal{F}(t) = \lambda \cdot \tanh\left(\frac{t}{100}\right)$, with the hyperparameter $\lambda$ set to $0.2$. Policy networks are implemented as 4-layer MLPs with a hidden width of 256. Training is conducted using a batch size of $256$, a learning rate of $3 \times 10^{-4}$, and the Adam optimizer~\citep{kingma2014adam}. For experiments involving selected subsets, data is repeated to match the size of the original dataset, and training is run for $20$ epochs to ensure equal training time for fair comparison.

\vspace{1mm}
\noindent \textbf{Metric \ \ } 
To evaluate the selected subset, we train policies on the subset with %standard 
BC, and obtain the averaged return by interacting for $10$ episodes. We use \texttt{\small{normalized return}} %\citep{fu2020d4rl} 
for better visualization: $\texttt{\small{normalized return}}=100 \times \frac{\texttt{return - random return}}{\texttt{expert return - random return}}$, where \texttt{\small{random return}} and \texttt{\small{expert return}} denote expected returns of random policies and the expert policy (online SAC \citep{haarnoja2018soft}), respectively.

%\paragraph{Baselines} 
\vspace{1mm}
%\noindent \textbf{Baselines \ \ } The most naive baseline is random selection (Random). Besides, we also employ the reward and Q-value as criteria for offline behavioral data selection. Three widely used coreset selection algorithms of Herding \citep{welling2009herding}, GraNd \citep{paul2021deep}, and GradMatch \citep{killamsetty2021grad} are also incorporated. These algorithms represent different categories of coreset selection approaches, providing a comprehensive benchmark for evaluation.

\noindent \textbf{Baselines\ \ } We compare against several representative baselines for offline behavioral data selection. The simplest is random sampling (Random). Additionally, we employ selection criteria based on reward (Top Reward) and Q-value (Top Q-value). To further strengthen our evaluation, we incorporate three widely adopted coreset selection algorithms: {Herding}~\citep{welling2009herding}, {GraNd}~\citep{paul2021deep}, and {GradMatch}~\citep{killamsetty2021grad}, each representing a distinct category of coreset approaches\footnote{We adopt the implementations from DeepCore~\citep{guo2022deepcore}, a widely used coreset selection library, using default hyperparameters.}.

%Following your advice, we selected a small subset of data with the same budget of 256 based on reward (Top Reward) and Q-value (Top Q-value). The results, presented in the table below, show that both Top Reward and Top Q-value perform significantly worse than random selection, indicating that simply filtering behavioral data based on reward or Q-value may not be effective. A possible explanation is that selecting data based on a single criterion, such as reward or Q-value, overlooks important factors like state diversity and long-horizon returns. In contrast, our OBD algorithm, SDW, effectively addresses these issues and synthesizes a compact yet informative behavioral dataset.

\begin{figure}[t]
    \centering
    		\includegraphics[width=0.325\columnwidth]{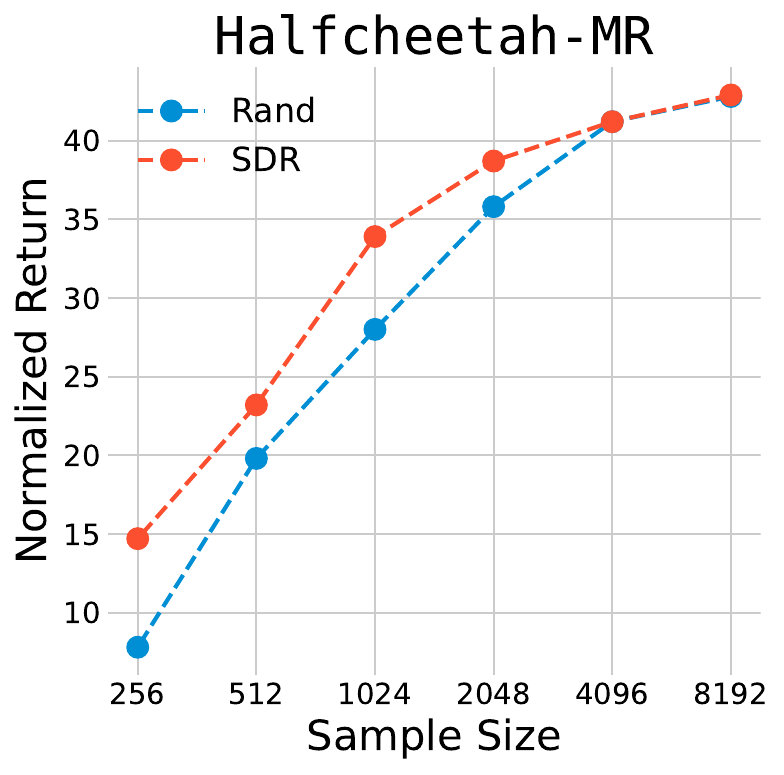}
            %\hspace{0.05\linewidth}
                \includegraphics[width=0.325\columnwidth]{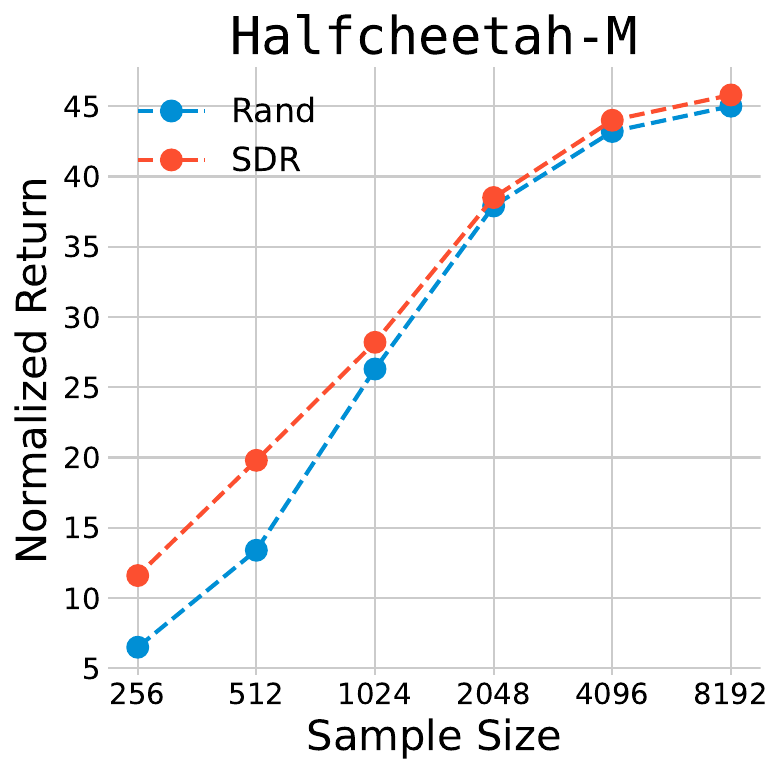}
                %\hspace{0.05\linewidth}
    		\includegraphics[width=0.325\columnwidth]{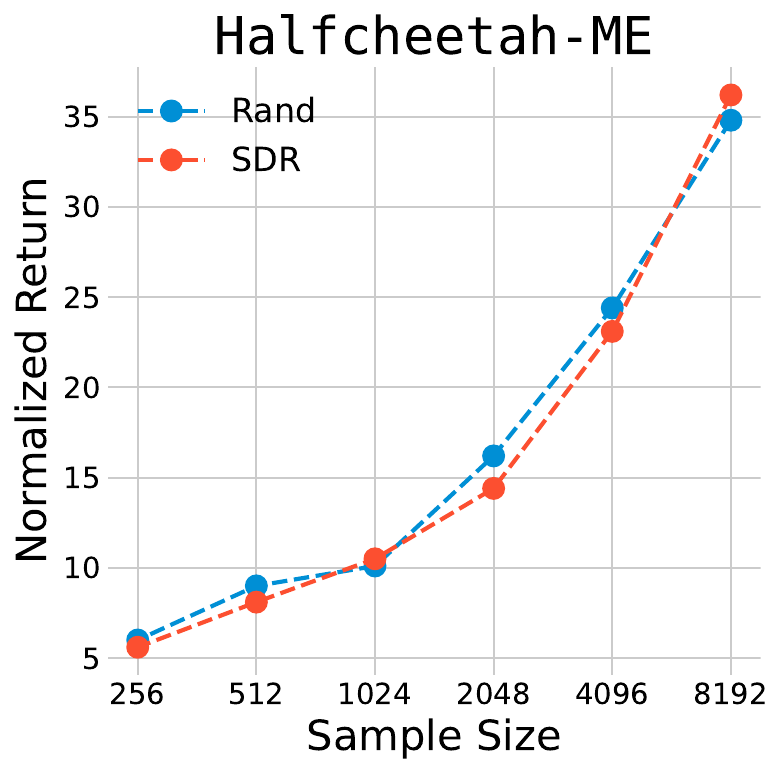}

                \includegraphics[width=0.325\columnwidth]{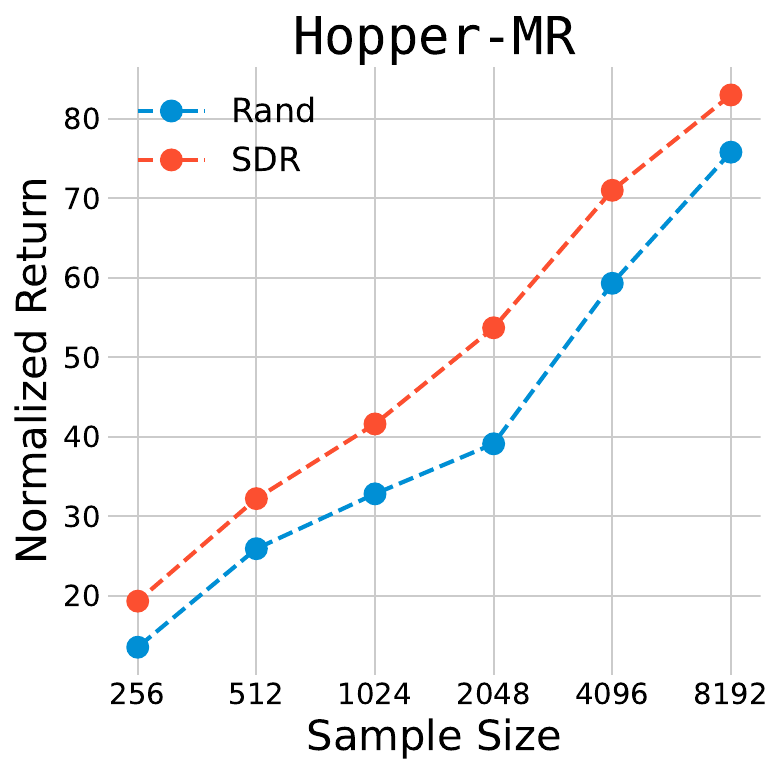}
                %\hspace{0.05\linewidth}
                \includegraphics[width=0.325\columnwidth]{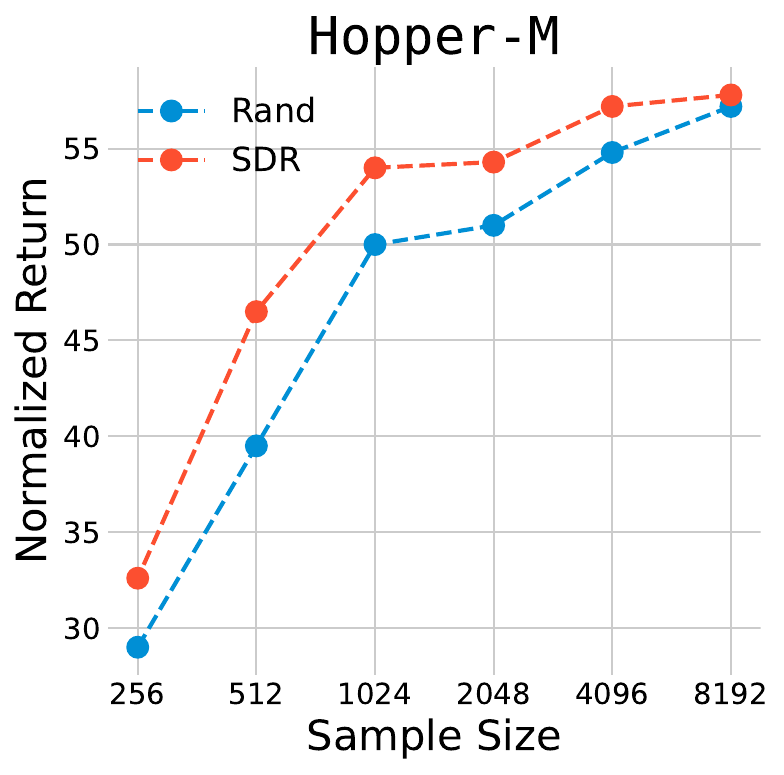}
                %\hspace{0.05\linewidth}
    		\includegraphics[width=0.325\columnwidth]{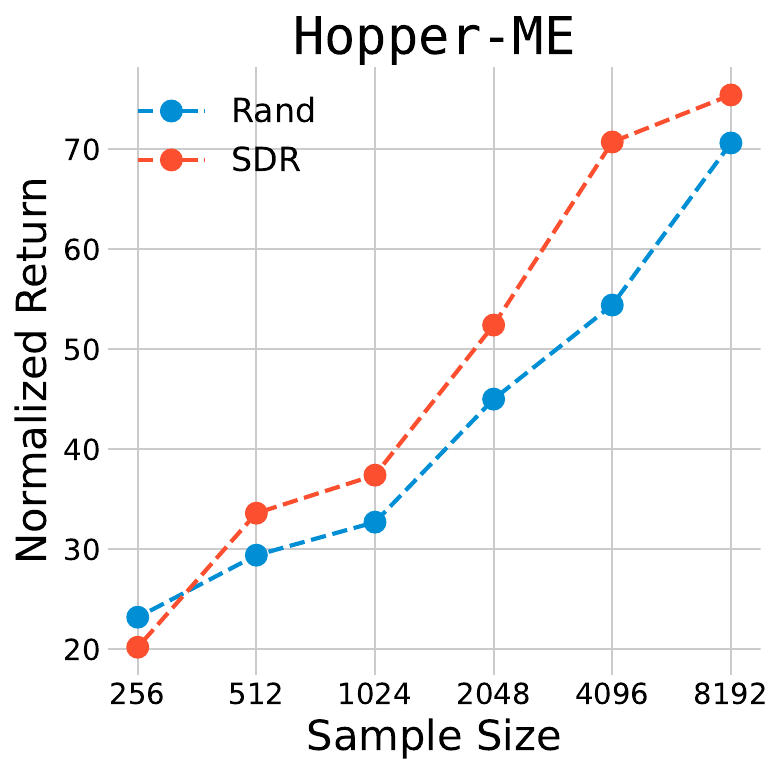}

                \includegraphics[width=0.325\columnwidth]{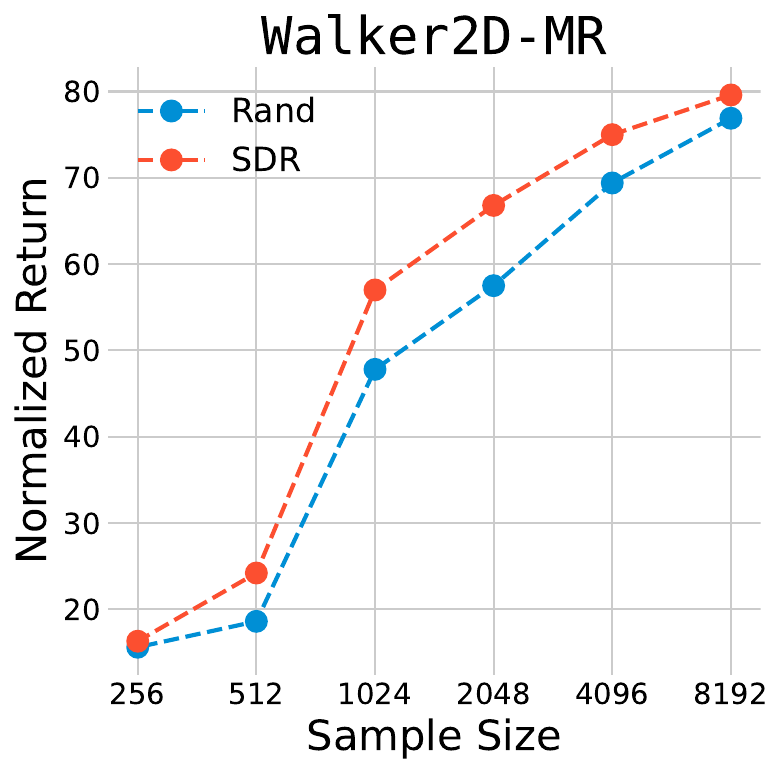}
                %\hspace{0.05\linewidth}
                \includegraphics[width=0.325\columnwidth]{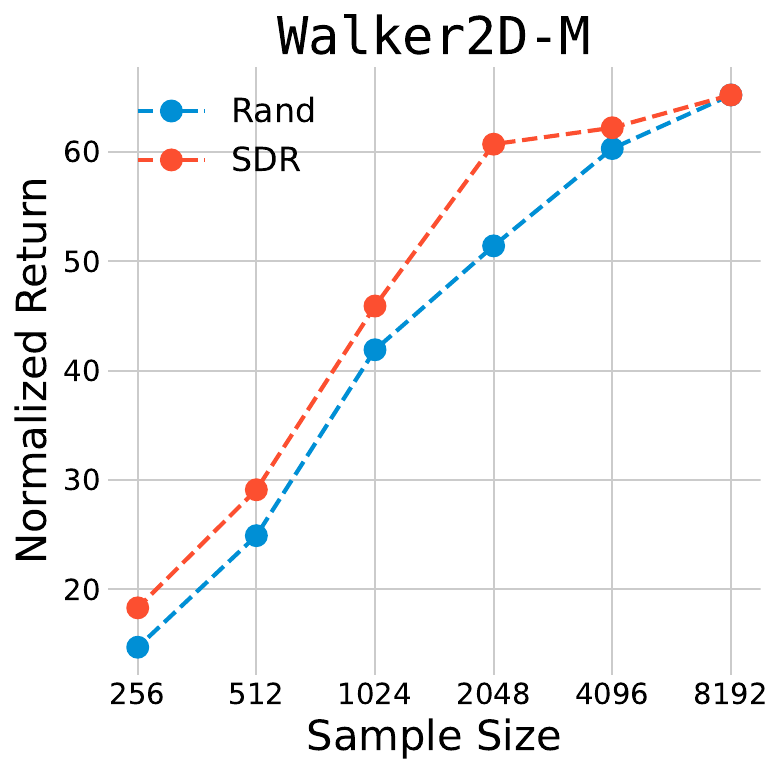}
                %\hspace{0.05\linewidth}
    		\includegraphics[width=0.325\columnwidth]{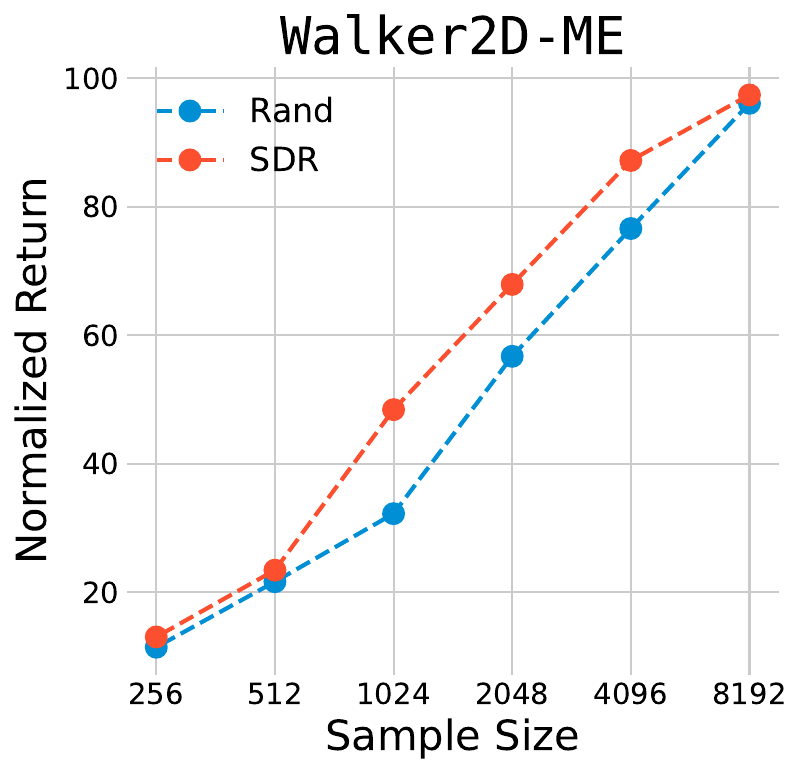}
    %\vspace{-1mm}
    \caption{Comparison of SDR and Random Selection (Rand) under different sample size. Each point is calculated and then averaged over ten trials.}
    \label{figure:multi_sample}  
\end{figure}

\subsection{Main Results}

We begin by evaluating the performance of various selection methods across datasets with varying quality levels and environments, as summarized in Table~\ref{tab:selection result}. Several key observations emerge from the results: (1) both {Top Reward} and {Top Q-value} perform significantly worse than the naive {Random} baseline. This suggests that relying on a single selection criterion, such as reward or Q-value, can be detrimental, likely because it overlooks other important factors such as diversity; (2) {Herding} and {GraNd} exhibit particularly poor performance in the context of offline behavioral data selection; (3) all three conventional coreset selection methods, namely {Herding}, {GraNd}, and {GradMatch}, underperform compared to the simple {Random} baseline; and (4) our proposed method, {SDR}, substantially improves offline data selection, achieving a notable performance gain over all baselines (from $27.0$ to $40.8$).

% We first evaluate the performance of various coreset selection methods across datasets of different quality levels and environments, as presented in Table \ref{tab:selection result}. Several key observations can be drawn from the results: (1) both Top Reward and Top Q-value have remarkable worse performance compared to naive random selection, a possible explanation is that selecting data based on a single criterion, such as reward or Q-value, overlooks other important factors like diversity; (2) Herding and GraNd exhibit notably poor performance in selecting offline behavioral data; (2) all three conventional coreset selection methods, Herding, GraNd, and GradMatch, perform worse than the simple random selection baseline; and (3) our proposed SDR significantly enhances offline behavioral data selection, achieving a substantial performance improvement over the baselines (from $27.0$ to $40.8$).

% 解释为什么传统方法不好
% 当loss比较大的时候，performance会迅速降低，所以geometry-based 方法不太work 

% geomory, loss, gradmatch

\vspace{1mm}

\noindent \textbf{SDR under various budgets \ \ }
%\paragraph{SDR under various budgets} 
Apart from evaluating SDR with a fixed budget of $1024$, we also assess its performance under different budget $N$ of $256$, $512$, $2048$, $4096$, and $8192$. Given the previously discussed suboptimal performance of conventional coreset selection methods, we primarily compare SDR against random selection. The results, presented in Figure \ref{figure:multi_sample}, reveal the following key observations: (1) SDR performs comparably to random selection in \texttt{HalfCheetah-ME} environment; and (2) for all other datasets, SDR consistently outperforms random selection across various budget sizes, further demonstrating its robustness and effectiveness.

\begin{table*}[t]
\scriptsize
    \centering
    \caption{Average return across different budgets on D4RL offline datasets. The best selection result for each dataset is marked with bold scores.}
    %\vspace{-0.1cm}
    \resizebox{1\linewidth}{!}{
    \begin{tabular}{c  ccc  ccc  ccc c}
    \toprule
     \multirow{2}{*}{Method}  & \multicolumn{3}{c}{Halfcheetach} & \multicolumn{3}{c}{Hopper} & \multicolumn{3}{c}{Walker2D} & \multirow{2}{*}{Average} \\
     & M-R & M & M-E & M-R & M & M-E & M-R & M & M-E \\
     %\specialrule{1pt}{0.2\jot}{0.2pc}
     \midrule
     Random & $29.2$ & $28.7$ & $\bm{16.8}$ & $41.1$ & $46.9$ & $42.6$ & $47.6$ &$43.1$ & $49.1$ & $38.3$   \\
     DualRank & $32.1$ & $28.9$ & $16.1$ & $48.1$ & $48.5$ & $43.2$ & $52.0$ &$45.8$ & $52.1$ & $41.8$   \\
     StepClip & $30.4$ & $29.8$ & $15.2$ & $46.9$ & $48.6$ & $43.5$ & $48.9$ &$45.9$ & $51.6$ & $40.1$   \\
     SDR & $\bm{32.4}$ & $\bm{31.3}$ & ${16.3}$ & $\bm{50.1}$ & $\bm{50.4}$ & $\bm{48.3}$ & $\bm{53.2}$ &$\bm{46.9}$ & $\bm{56.2}$ & $\bm{42.8}$   \\
    \bottomrule
    \end{tabular}
    }
    %\vspace{-0.6cm}
    \label{tab:ablation result}
\end{table*}

\begin{figure}[t]
    \centering
    		\includegraphics[width=0.325\columnwidth]{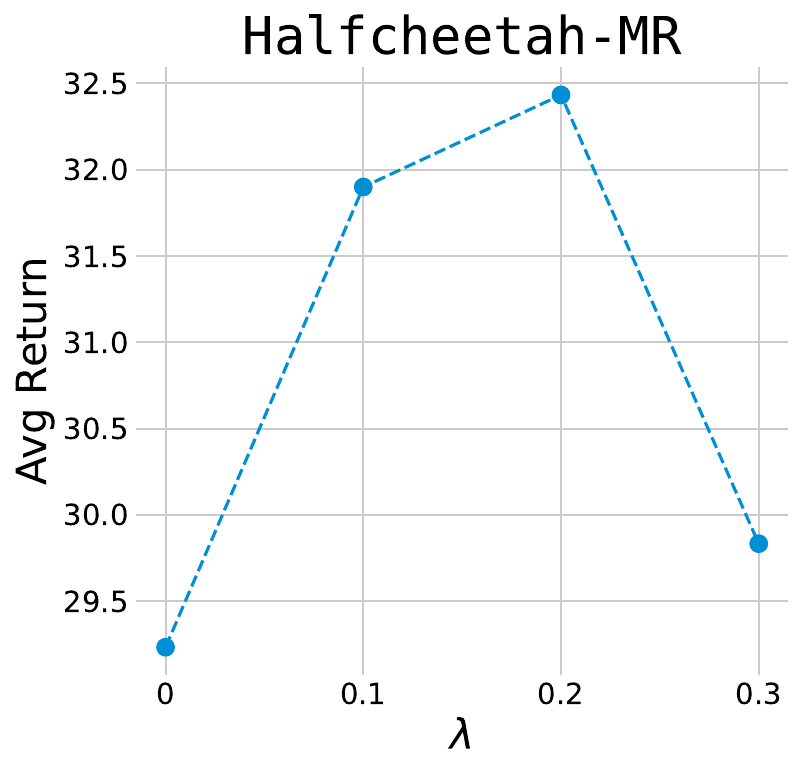}
                \includegraphics[width=0.325\columnwidth]{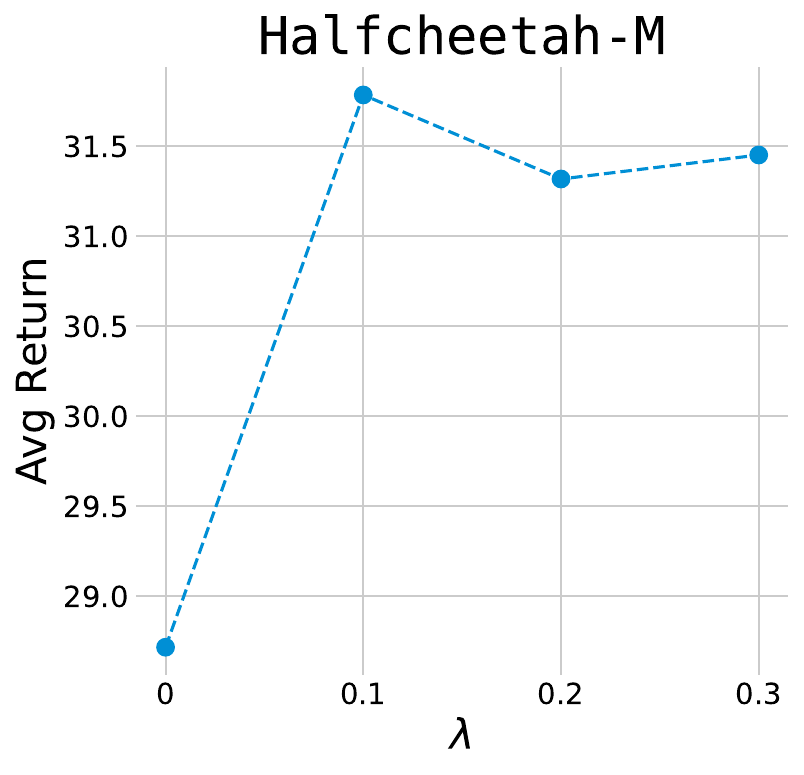}
    		\includegraphics[width=0.325\columnwidth]{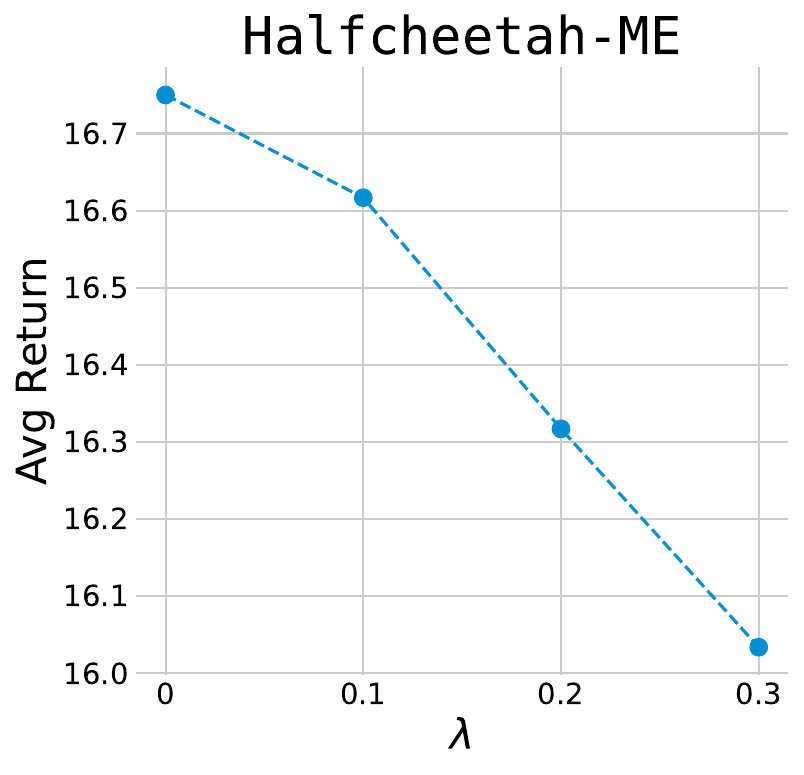}

                \includegraphics[width=0.325\columnwidth]{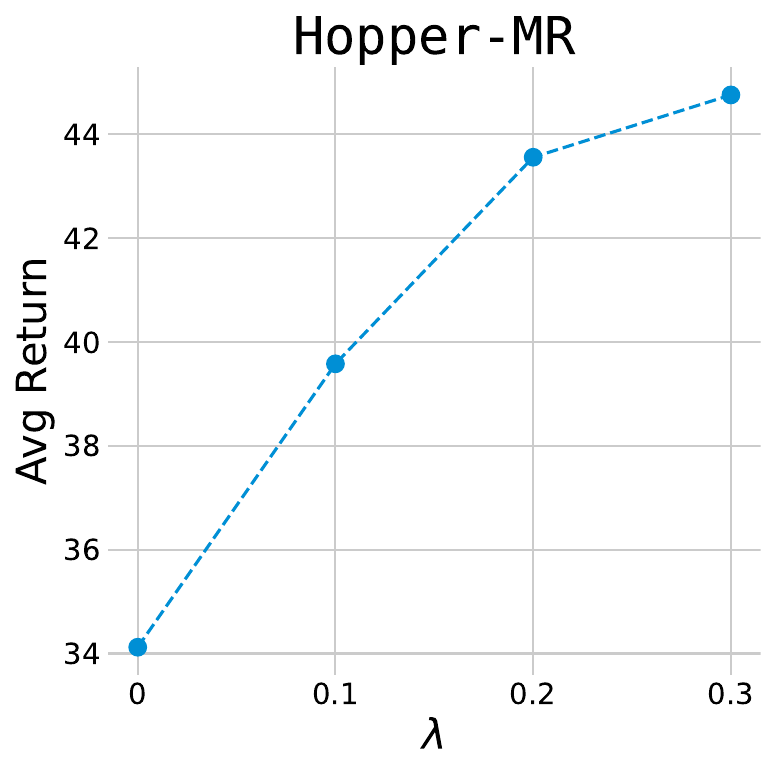}
                \includegraphics[width=0.325\columnwidth]{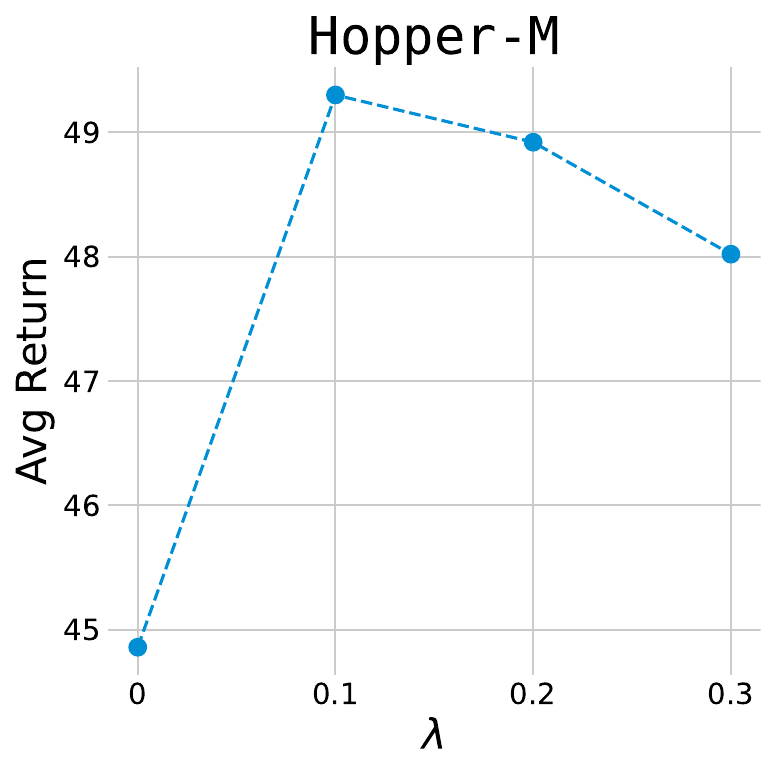}
    		\includegraphics[width=0.325\columnwidth]{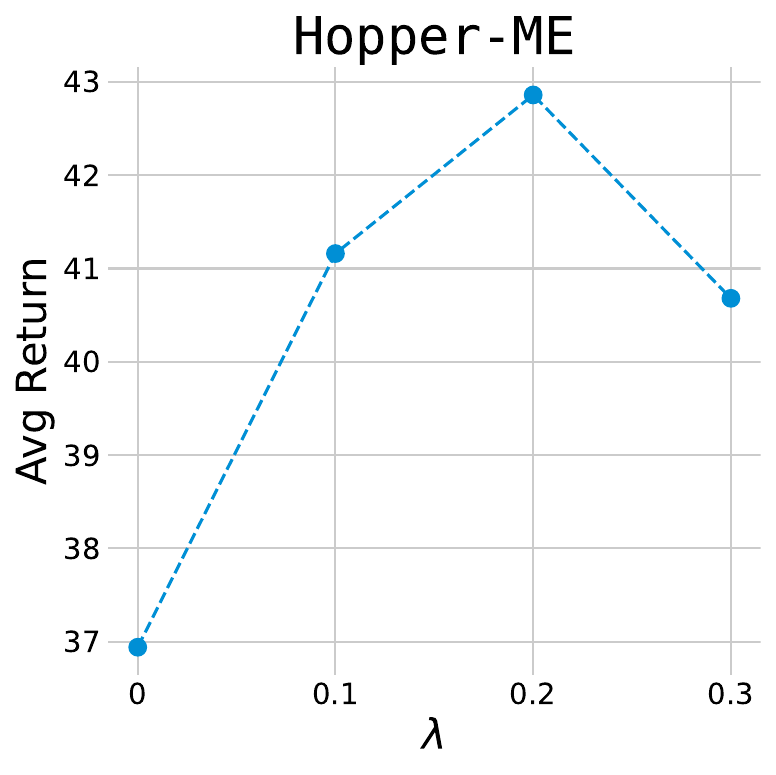}

    		\includegraphics[width=0.325\columnwidth]{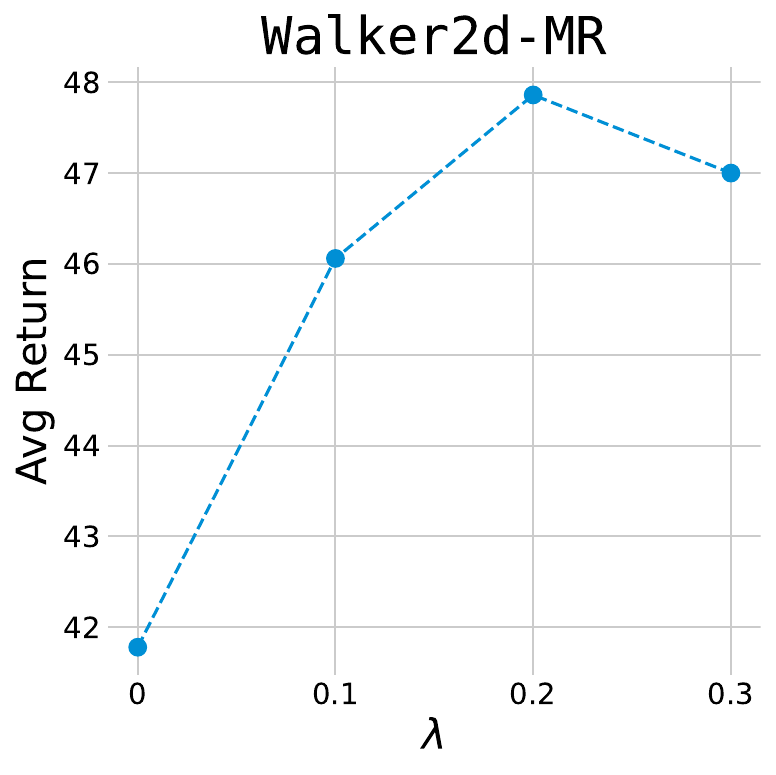}
                \includegraphics[width=0.325\columnwidth]{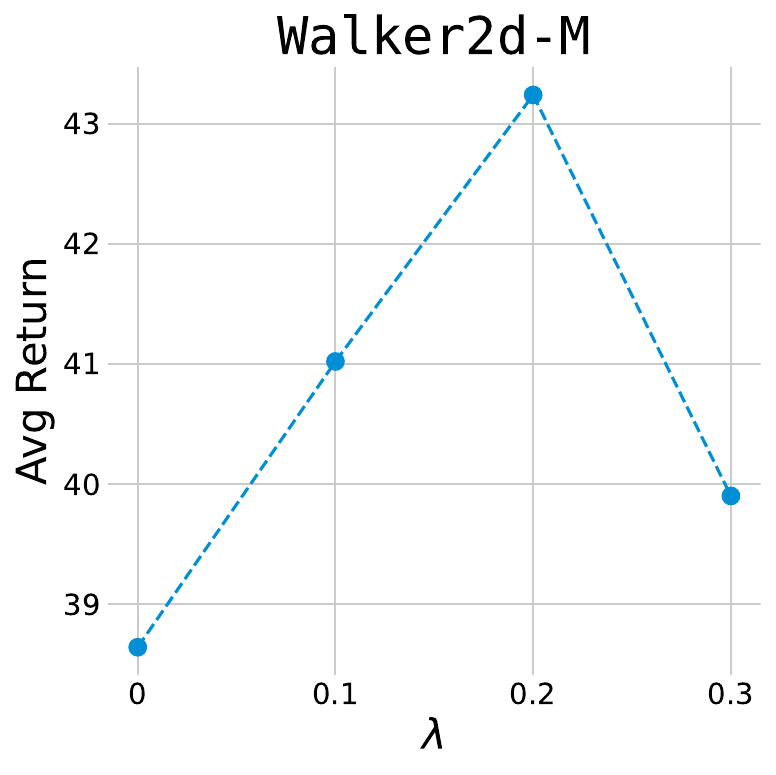}
    		\includegraphics[width=0.325\columnwidth]{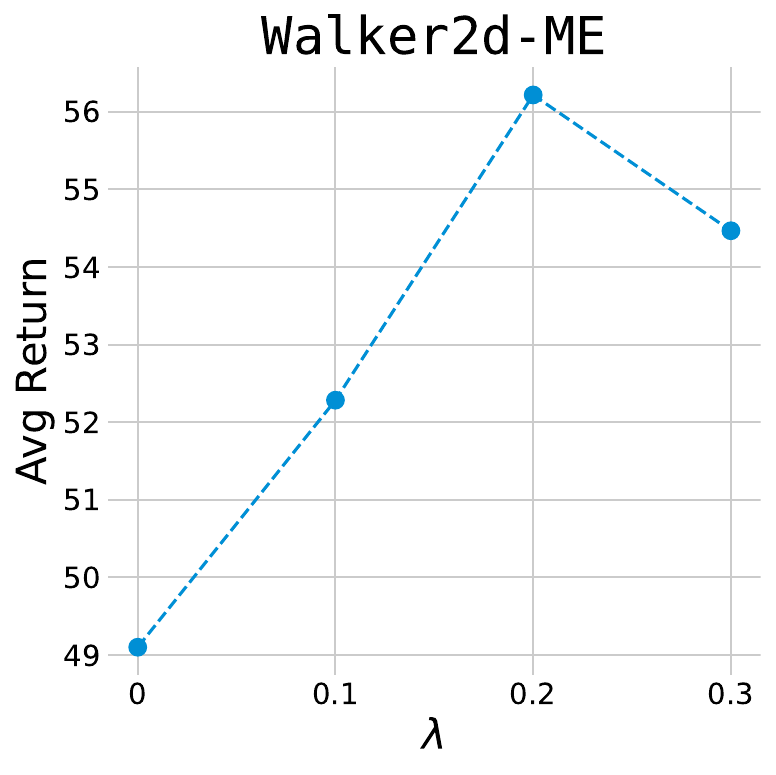}
    %\vspace{-1mm}
    \caption{Plots of policy performance as functions of $\lambda$. Each point is calculated and then averaged over ten trials.}
    \label{figure:lambda}   
\end{figure}

\subsection{Ablation Study}
As outlined in Section \ref{sec:algorithm}, SDR comprises two key components: (1) Stepwise Progressive Clip and (2) Dual Ranking. In this section, we conduct ablation experiments to evaluate their effectiveness.

To assess the impact of Stepwise Progressive Clip, we replace the monotonic function $\mathcal{F}(t) = \lambda \tanh{\frac{t}{100}}$ with a constant function $\mathcal{F}(t) = \lambda$, while keeping Dual Ranking unchanged. Conversely, to examine the role of Dual Ranking, we replace it with a single-ranking strategy that selects examples based solely on the ratio $q_{\pi^\ast}(s,a)/d_{\pi_\beta}(s)$. We evaluate the average return under different budget sizes of $[256, 512, 1024, 2048, 4096, 8192]$, with results presented in Table \ref{tab:ablation result}. From the results, we observe that: (1) employing either Stepwise Progressive Clip or Dual Ranking individually improves selection performance compared to the baseline random selection; and (2) integrating both components in SDR further enhances performance, verifying the effectiveness of SDR’s design.

% 之间sample出specific number的xxx

\subsection{Hyperparameter Analysis}
\label{sec:hyperparameter}
% hyperparameter的meaning是什么
% 不同hyperparameter对于SDR在不同dataset上的影响

% lambda = 0, 0.1, 0.2, 0.3 (ours)
The hyperparameter $\lambda$ in $\mathcal{F}(t)$ controls the selection intensity and the size of candidate pool. A smaller $\lambda$ reduces the sensitivity to the rankings induced by $q_{\pi^\ast}$ and $d_{\pi_\beta}$, thereby permitting lower-ranked examples to be included and increasing the overall size of the candidate pool. To investigate the impact of $\lambda$, we evaluate SDR with varying values of $\lambda$ of $0.1$, $0.2$ and $0.3$. The average return across different budgets, as presented in Figure \ref{figure:lambda}, reveals that: (1) $\lambda$ influence SDR performance; (2) SDR with $\lambda$ values of $0.1, 0.2$, and $0.3$ consistently outperforms the baseline, except in \texttt{Halfcheetah-ME}; and (3) different datasets exhibit distinct optimal $\lambda$ values. We also provide insights into selecting an appropriate $\lambda$. For offline behavioral data collected from suboptimal policies, a larger $\lambda$ should be employed to better correct the distribution shift between behavior and expert policies.

\section{Discussion and Limitation}
\label{sec:discussion}
\noindent \textbf{Discussion \ \ }
%\paragraph{Discussion} 
The primary strength of SDR lies in its efficiency, a crucial property for data selection algorithms aiming to enable efficient training. Many dataset compression methods require repeated training over the full dataset or expensive second-order computations (e.g., Hessian matrices), resulting in overheads that can be several orders of magnitude greater than simply training on the full dataset. In contrast, once estimates of the action value and state density are obtained, SDR performs data selection without any additional training, making it entirely training-free and highly scalable. This efficiency allows SDR to scale seamlessly to large offline behavioral datasets, which is increasingly important in the era of large-scale data. Moreover, SDR allows flexible subset sizing by simply adjusting the number of samples drawn from the candidate pool, with no recomputation required. This further enhances its practicality and efficiency.

\vspace{2mm}

\noindent \textbf{Limitation \ \ }
%\paragraph{Limitation} 
The current SDR algorithm has two limitations. First, it requires pre-estimation of both the action value function and the density function. While the density function can be estimated solely from the states in the original dataset, estimating the action value function requires access to a reward-labeled RL dataset. As a result, SDR necessitates an offline RL dataset composed of full transition tuples including rewards. Second, %as demonstrated in Section \ref{sec:hyperparameter}, 
the performance of SDR is influenced by the hyperparameter $\lambda$, which controls the trade-off between diversity and importance. Developing principled methods for automatically tuning $\lambda$ remains an open and promising direction for future work.

% designed for non-expert sampled offline dataset 

\section{Conclusion}

This paper identifies and analyzes a critical inefficiency in behavioral cloning: data saturation, where policy performance quickly plateaus despite increasing the size of training data. We theoretically attribute this phenomenon to the weak alignment between test loss and true policy performance under distributional shift. Motivated by this insight, we introduce \textit{Stepwise Dual Ranking} (SDR), a simple yet effective method for selecting a compact and informative subset from large-scale offline behavioral datasets. SDR combines stepwise clipping, which emphasizes early-timestep data, with a dual-ranking strategy that favors high-value, low-density states. Empirical results across a range of D4RL benchmarks show that SDR substantially improves data efficiency, enabling strong policy performance with significantly fewer training samples.

\noindent \textbf{Acknowledgment \ \ }
Dr Tao’s research is partially supported by NTU RSR and Start Up Grants.

%In this paper, we investigate the issue of data saturation in behavioral cloning (BC) and reveal that policy performance does not necessarily improve as training loss decreases. Through theoretical analysis, we demonstrate that this phenomenon arises due to the state distribution shift between the expert policy and the behavior policy. To address this issue, we propose Selective Data Reduction (SDR), a novel and computationally efficient method for constructing a compact yet informative subset from large-scale offline behavioral datasets. SDR employs a stepwise selection mechanism that prioritizes high-policy-value and low-state-density samples, effectively mitigating data saturation. Extensive experiments and ablation studies on multiple D4RL datasets validate the effectiveness of SDR in reducing dataset size while maintaining or even improving policy performance.

\balance
\bibliographystyle{ACM-Reference-Format}
\bibliography{ref}

\appendix

\section{Proof of Lemma \ref{lemma:generalization}}
\label{app:lemma}
\begin{proof}

For each example $i \in \{1, \dots, m\}$, the contribution to the empirical error is $Z_i = \frac{w_i}{m} D_{\operatorname{TV}}(\pi, \pi^\ast)$, where $w_i = \frac{d(s_i)}{d{\prime}(s_i)}$. Since $D_{\operatorname{TV}}(\pi, \pi^\ast) \in [0,1]$, we have $Z_i \in [0, \frac{w_i}{m}]$. Applying Hoeffding’s inequality for bounded independent random variables, with total range at most $\frac{w_i}{m}$ per term, we obtain that, with probability at least $1 - \delta$,
\begin{equation}
    \epsilon \le \hat{\epsilon} + \sqrt{ \frac{ \sum_{i=1}^m w_i^2 \log \frac{2}{\delta} }{m^2} }.
\end{equation}
This establishes the claimed upper bound on the expected error.
\end{proof}

\iffalse
\section{Implementation Details}
\label{app:implementation details}
This section provides all the additional implementation details of our experiments.

\noindent \textbf{Poilcy Training \ \ }
%\paragraph{Policy Training} 
We use a 4-layer multilayer perceptron (MLP) with a hidden width of $256$ as the policy network. Training is performed with a batch size of $256$, a learning rate of $3 \times 10^{-4}$, and the Adam optimizer \citep{kingma2014adam}. For experiments using selected subsets, we repeat the data to match the size of the original dataset and train for $20$ epochs, ensuring equal training time for a fair comparison.

\vspace{1mm}
\noindent \textbf{Other Selection Methods \ \ }
%\paragraph{Other Selection Methods} 
For the coreset selection baselines reported in Table \ref{tab:selection result}, we adopt the implementation from DeepCore \citep{guo2022deepcore}, a widely used coreset selection library, and use its default hyperparameters.
\fi

\end{document}